\documentclass[journal]{IEEEtran}
\usepackage{cite}
\usepackage{amsmath,graphicx, multirow}
\usepackage{graphicx}
\usepackage{subfig}
\usepackage{flushend}
\usepackage{float}
\usepackage{mathtools}
\usepackage{color}
\usepackage{enumerate}
\usepackage{bm}
\usepackage{multirow}
\usepackage{array}
\usepackage[utf8]{inputenc}
\usepackage{amssymb}
\usepackage{makecell}
\usepackage{booktabs}
\usepackage{diagbox}
\def\BibTeX{{\rm B\kern-.05em{\sc i\kern-.025em b}\kern-.08em
    T\kern-.1667em\lower.7ex\hbox{E}\kern-.125emX}}
\usepackage{balance}

\begin{document}

\title{SC$^{3}$EF: A Joint Self-Correlation and Cross-Correspondence Estimation Framework for Visible and Thermal Image Registration}

\author{
	
	{Xi Tong,
	Xing Luo,  
	Jiangxin Yang, 
	Xin Li,~\IEEEmembership{Senior Member,~IEEE},
	and Yanpeng Cao,~\IEEEmembership{Senior Member,~IEEE}
    }
	
	\thanks{This work was supported by the National Natural Science Foundation of China under Grant 52075485, 52275548, and the Aeronautical Science Foundation of China under Grant 2022Z071076002. (Corresponding author: Yanpeng Cao.)
	
	X. Tong, X. Luo, J. Yang, and Y. Cao are with the State Key Laboratory of Fluid Power and Mechatronic Systems and Laboratory of Advanced Manufacturing Technology of Zhejiang Province, School of Mechanical Engineering, Zhejiang University, Hangzhou 310027, China. 
		
	X. Li is with Section of Visual Computing and Creative Media, School of Performance, Visualization, and Fine Arts, Texas A\&M University, College Station, Texas 77843, United States of America.}}

\markboth{Journal of \LaTeX\ Class Files,~Vol.~18, No.~9, September~2020}%
{How to Use the IEEEtran \LaTeX \ Templates}

\maketitle

\begin{abstract}

Multispectral imaging plays a critical role in a range of intelligent transportation applications, including advanced driver assistance systems (ADAS), traffic monitoring, and night vision. \textcolor{black}{However, accurate visible and thermal (RGB-T) image registration poses a significant challenge due to the considerable modality differences.} In this paper, we present a novel joint Self-Correlation and Cross-Correspondence Estimation Framework (SC$^{3}$EF), leveraging both local representative features and global contextual cues to effectively generate RGB-T correspondences. For this purpose, we design a convolution-transformer-based pipeline to extract local representative features and encode global correlations of intra-modality for inter-modality correspondence estimation between unaligned visible and thermal images. After merging the local and global correspondence estimation results, we further employ a hierarchical optical flow estimation decoder to progressively refine the estimated dense correspondence maps. Extensive experiments demonstrate the effectiveness of our proposed method, outperforming the current state-of-the-art (SOTA) methods on representative RGB-T datasets. \textcolor{black}{Furthermore, it also shows competitive generalization capabilities across challenging scenarios, including large parallax, severe occlusions, adverse weather, and other cross-modal datasets (e.g., RGB-N and RGB-D).}

\end{abstract}


\section{Introduction}

\textcolor{black}{Multispectral imaging is essential for capturing complementary multi-modality information, such as RGB-Thermal (RGB-T) image pairs, to facilitate emerging applications in autonomous driving and traffic monitoring. A key factor in the success of these tasks is the acquisition of well-aligned visible and thermal image pairs. In advanced driver assistance systems (ADAS), aligned RGB-T imagery can enable improved pedestrian and obstacle detection, allowing self-driving vehicles to better perceive their surroundings in various lighting and weather conditions \cite{li2019rgb}. Similarly, in intelligent traffic management systems, the fusion of well-aligned RGB-T data can enhance capabilities such as vehicle tracking, traffic flow analysis, and incident detection, ultimately improving both road safety and efficiency \cite{xu2020u2fusion}. Most existing RGB-T imaging systems utilize a binocular vision design, utilizing separate visible and thermal cameras for image acquisition \cite{qiao2017integrative}. However, differences in spectral response and spatial configuration between RGB-T sensors will cause significant discrepancies in the captured visible and thermal images, making high-accuracy RGB-T image registration a long-standing challenge. More specifically, visible images capture light that is reflected in the visible spectrum, whereas thermal infrared images represent the radiation emitted by the target, known as thermal radiation which is highly correlated with the temperature of the object \cite{gade2014thermal}. According to the laws of thermodynamics, thermal radiation has diffusivity, allowing it to transfer from a high-temperature object to a low-temperature one.}


\begin{figure}[t]
	\normalsize
	\centering
	\includegraphics[scale=0.35]{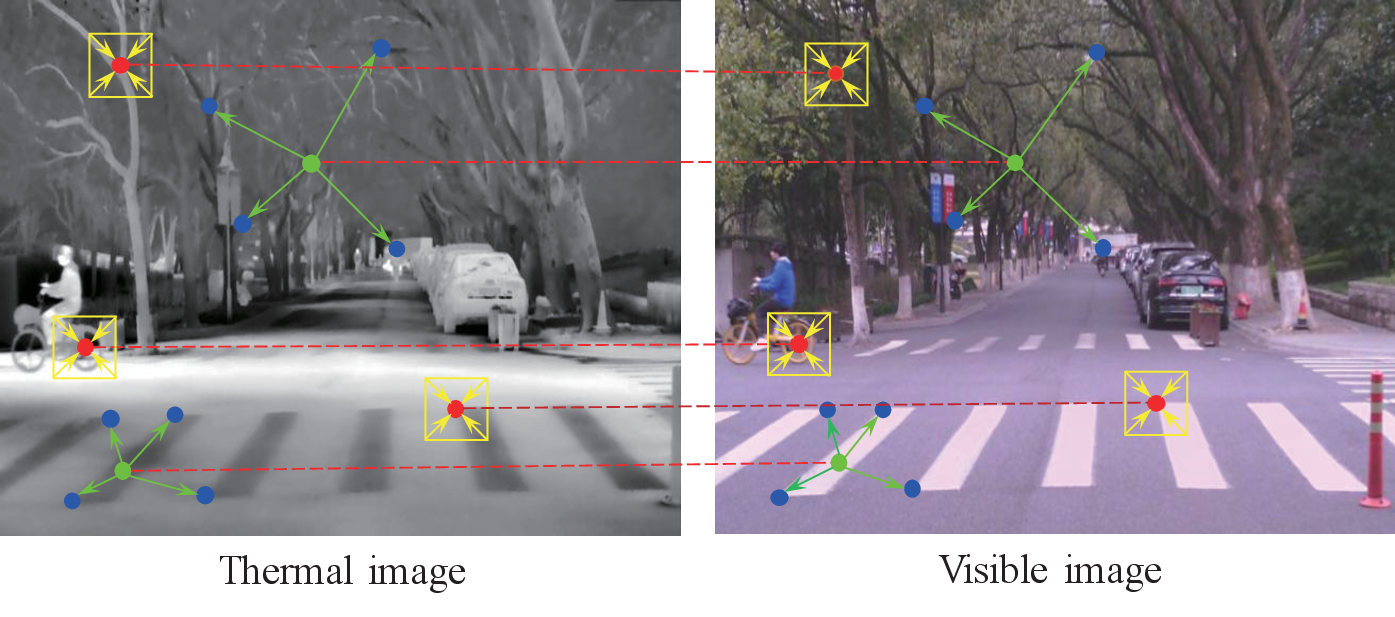}
	\caption{An illustration of the use of both local representative features and global contextual cues for human observers to accurately identify correspondences between unaligned visible and thermal images.}
	\label{scheme}
\end{figure}

\begin{figure*}[t]
	\normalsize
	\centering
	\includegraphics[scale=0.30]{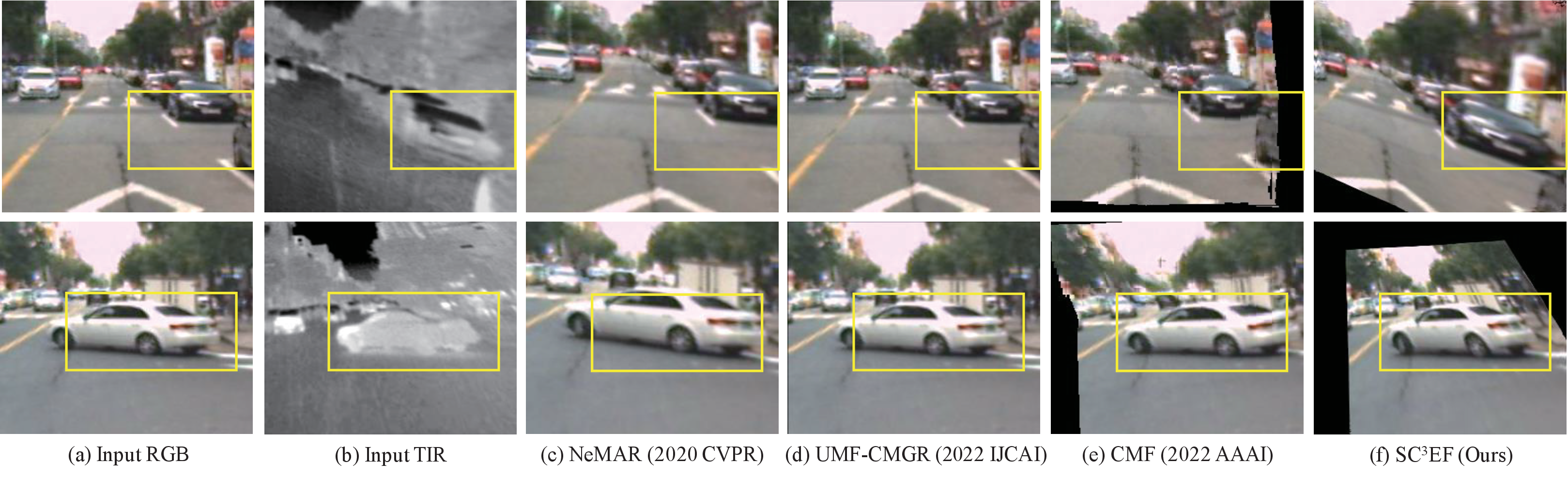}
	\caption{Comparative results of a number of SOTA cross-modality registration methods including NeMAR \cite{arar2020unsupervised}, UMF-CMGR \cite{wang2022unsupervised}, CMF \cite{zhou2022promoting} and our proposed SC$^{3}$EF method. It is observed that SC$^{3}$EF can achieve more accurate and robust registration of visible and thermal images with significant misalignment. Please zoom in to check the details highlighted in the yellow bounding boxes.}
	\label{intro_example}
\end{figure*}

\begin{table*}[tb]
	\Large
	\renewcommand{\arraystretch}{1.2}
	\centering
	\caption{\textcolor{black}{Overview of traditional and deep learning-based methods for mono-modal and cross-modal image registration.}}
	\resizebox{0.95\textwidth}{!}{
		\begin{tabular}{cccccccccc}
			\hline
			\multicolumn{1}{c}{Method} &
			\multicolumn{1}{c}{Category} & 
			\multicolumn{1}{c}{Type} &
			\multicolumn{1}{c}{Modality} &
			\multicolumn{1}{c}{Dataset/Scene} &
			\multicolumn{1}{c}{Warp type} &
			\multicolumn{1}{c}{Core idea} & 
			\multicolumn{1}{c}{Large parallax} &
			\multicolumn{1}{c}{Real-world val.} & 
			\multicolumn{1}{c}{Cross-weather} \\
			\hline
			\multicolumn{1}{c}{\cite{chen2017normalized}} &
			\multirow{4}{*}[-2.5ex]{Traditional} & 
			\multicolumn{1}{c}{Area-based} &
			\multicolumn{1}{c}{Multispectral} &
			\multicolumn{1}{c}{Multiple} &
			\multicolumn{1}{c}{Rigid} &
			\multicolumn{1}{c}{\makecell{Normalized total \\ gradient measure}} & 
			\multicolumn{1}{c}{$\times$} &
			\multicolumn{1}{c}{\checkmark} & 
			\multicolumn{1}{c}{$\times$} \\
			\multicolumn{1}{c}{\cite{cao2020boosting}} &
			~ & \multicolumn{1}{c}{Area-based} &
			\multicolumn{1}{c}{Multimodal} &
			\multicolumn{1}{c}{Multiple} &
			\multicolumn{1}{c}{Rigid} &
			\multicolumn{1}{c}{\makecell{Boosting structure \\ consistency}} & 
			\multicolumn{1}{c}{$\times$} &
			\multicolumn{1}{c}{\checkmark} & 
			\multicolumn{1}{c}{$\times$} \\
			\multicolumn{1}{c}{\cite{sonn2013fast}} &
			~ & \multicolumn{1}{c}{Feature-based} &
			\multicolumn{1}{c}{RGB-T} &
			\multicolumn{1}{c}{Person/Surveillance} &
			\multicolumn{1}{c}{Rigid} &
			\multicolumn{1}{c}{Noisy polygon vertices} & 
			\multicolumn{1}{c}{$\times$} &
			\multicolumn{1}{c}{\checkmark} & 
			\multicolumn{1}{c}{$\times$} \\
			\multicolumn{1}{c}{\cite{jiang2020contour}} &
			~ & \multicolumn{1}{c}{Feature-based} &
			\multicolumn{1}{c}{RGB-T} &
			\multicolumn{1}{c}{Power device} &
			\multicolumn{1}{c}{Rigid} &
			\multicolumn{1}{c}{\makecell{Curvature scale space \\ (CSS) corner detector}} & 
			\multicolumn{1}{c}{$\times$} &
			\multicolumn{1}{c}{\checkmark} & 
			\multicolumn{1}{c}{$\times$} \\
			\hline
			\multicolumn{1}{c}{\cite{rocco2017convolutional}} &
			\multirow{8}{*}[-8ex]{Learning} & 
			\multicolumn{1}{c}{Transform model} &
			\multicolumn{1}{c}{RGB} &
			\multicolumn{1}{c}{Multiple} &
			\multicolumn{1}{c}{Rigid} &
			\multicolumn{1}{c}{\makecell{CNN for keypoint \\ extraction}} & 
			\multicolumn{1}{c}{$\times$} &
			\multicolumn{1}{c}{\checkmark} & 
			\multicolumn{1}{c}{$\times$} \\
			\multicolumn{1}{c}{\cite{xu2022rfnet}} &
			~ & \multicolumn{1}{c}{Transform model} &
			\multicolumn{1}{c}{Multimodal} &
			\multicolumn{1}{c}{Traffic} &
			\multicolumn{1}{c}{Rigid} &
			\multicolumn{1}{c}{\makecell{Style transfer + \\ Deformable CNN}} & 
			\multicolumn{1}{c}{$\times$} &
			\multicolumn{1}{c}{$\times$} & 
			\multicolumn{1}{c}{$\times$} \\
			\multicolumn{1}{c}{\cite{melekhov2019dgc}} &
			~ & \multicolumn{1}{c}{Deformation} &
			\multicolumn{1}{c}{RGB} &
			\multicolumn{1}{c}{Multiple} &
			\multicolumn{1}{c}{Non-rigid} &
			\multicolumn{1}{c}{Coarse-to-fine CNN} & 
			\multicolumn{1}{c}{$\times$} &
			\multicolumn{1}{c}{\checkmark} & 
			\multicolumn{1}{c}{$\times$} \\
			\multicolumn{1}{c}{\cite{xu2022gmflow}} &
			~ & \multicolumn{1}{c}{Deformation} &
			\multicolumn{1}{c}{RGB} &
			\multicolumn{1}{c}{Animation/Traffic} &
			\multicolumn{1}{c}{Non-rigid} &
			\multicolumn{1}{c}{\makecell{Transformer + Global \\ correlation}} & 
			\multicolumn{1}{c}{\checkmark} &
			\multicolumn{1}{c}{\checkmark} & 
			\multicolumn{1}{c}{$\times$} \\
			\multicolumn{1}{c}{\cite{arar2020unsupervised}} &
			~ & \multicolumn{1}{c}{Deformation} &
			\multicolumn{1}{c}{Multimodal} &
			\multicolumn{1}{c}{Plants} &
			\multicolumn{1}{c}{Non-rigid} &
			\multicolumn{1}{c}{\makecell{Style transfer for self- \\ supervision}} & 
			\multicolumn{1}{c}{$\times$} &
			\multicolumn{1}{c}{\checkmark} & 
			\multicolumn{1}{c}{$\times$} \\
			\multicolumn{1}{c}{\cite{zhou2022promoting}} &
			~ & \multicolumn{1}{c}{Deformation} &
			\multicolumn{1}{c}{Multimodal} &
			\multicolumn{1}{c}{Traffic} &
			\multicolumn{1}{c}{Non-rigid} &
			\multicolumn{1}{c}{\makecell{Single-modal teacher \\ network}} & 
			\multicolumn{1}{c}{$\times$} &
			\multicolumn{1}{c}{\checkmark} & 
			\multicolumn{1}{c}{$\times$} \\
			\multicolumn{1}{c}{\cite{wang2022unsupervised}} &
			~ & \multicolumn{1}{c}{Deformation} &
			\multicolumn{1}{c}{RGB-T} &
			\multicolumn{1}{c}{Traffic} &
			\multicolumn{1}{c}{Non-rigid} &
			\multicolumn{1}{c}{\makecell{Style transfer + CNN- \\ based flow estimation}} & 
			\multicolumn{1}{c}{$\times$} &
			\multicolumn{1}{c}{$\times$} & 
			\multicolumn{1}{c}{$\times$} \\
			\multicolumn{1}{c}{Ours} &
			~ & \multicolumn{1}{c}{Deformation} &
			\multicolumn{1}{c}{RGB-T} &
			\multicolumn{1}{c}{Traffic} &
			\multicolumn{1}{c}{Non-rigid} &
			\multicolumn{1}{c}{\makecell{Frequency-specific CNN-\\ Transformer + Self- and cross- \\ correspondences}} & 
			\multicolumn{1}{c}{\checkmark} &
			\multicolumn{1}{c}{\checkmark} & 
			\multicolumn{1}{c}{\checkmark} \\
			\hline
	\end{tabular}}
	\label{related_tab}
\end{table*}

\textcolor{black}{Recent advancements in correspondence estimation techniques for unaligned visible images have led to a surge in research on cross-modal image registration, including RGB-T \cite{wang2022unsupervised}, RGB-N \cite{han2023cross}, and RGB-D \cite{zhou2022promoting}. The existing methods primarily utilize hand-crafted descriptors \cite{cao2020boosting, chen2010partial, jiang2020contour} or learning-based optical flow estimation \cite{wang2022unsupervised, zhou2022promoting, xu2022rfnet, han2023cross}.} Descriptor-based approaches rely on geometric feature extraction (e.g., corners, gradients, contours) but struggle with content variations in visible and thermal image pairs, leading to inaccurate results \cite{cui2021cross}. In comparison, the learning-based optical flow estimation methods typically utilize a style transfer sub-network as the pre-processing to bridge the \textcolor{black}{differences} between modalities and deploy a convolutional sub-network to extract distinctive local features for \textcolor{black}{cross-modal} image matching. \textcolor{black}{However, these methods require prior knowledge of modal characteristics in specific scenes, and the pseudo-thermal images converted from visible ones do not always appear realistic when compared to real-captured thermal images.} Additionally, the commonly-used convolutional backbones cannot generate satisfactory results when performing large displacement registration tasks due to the fixed window size of matching cost volume \cite{xu2022gmflow}.

\begin{figure*}[t]
	\normalsize
	\centering
	\includegraphics[scale=0.37]{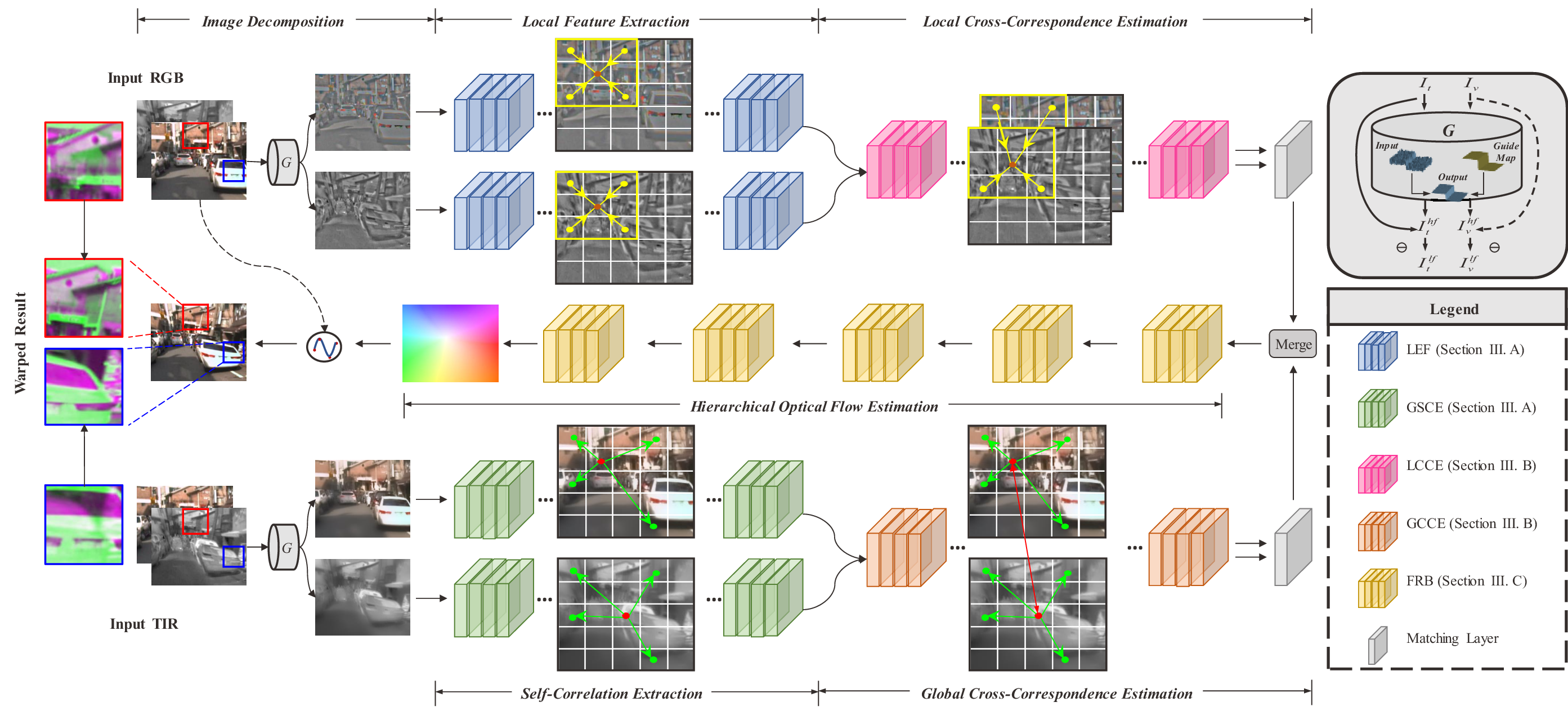}
	\caption{\textcolor{black}{The overview of our proposed SC$^{3}$EF for RGB-T image registration, which contains four main stages including image decomposition, local feature \& self-correlation extraction, local- \& global- correspondence estimation, and hierarchical optical flow estimation. Key steps and registration performance (before and after) are zoomed in for clarity.}}
	\label{overall_arch}
\end{figure*}


\textbf{The inspiration for this research work is drawn from how human observers can accurately establish correspondences between significantly different visible and thermal images.} Human observers tend to rely on local representations to identify correspondences for regions with enough mutual details or textures, while utilizing global contextual information to establish correspondences in regions lacking distinct and mutual features, as illustrated in Fig.~\ref{scheme}. For instance, it is feasible to establish correspondences in visible and thermal images by constructing distinctive and mutual features based on textures and edges in local regions (e.g., edges of bicycle wheels, textures of tree branches, or corners of zebra-crossings). However, visible and thermal images typically exhibit significantly different characteristics, and thus it is difficult to gather enough mutual features to facilitate robust multispectral image registration. As an effective remedy, human observers will perceive the global configuration of relevant objects or features within one modality and utilize this important self-correlation cue to search for correspondences across modalities. \textbf{Therefore, it is important to make use of both local representative features and global contextual cues to identify correspondences between unaligned visible and thermal images.
}

Based on the above considerations, in this paper, we present a novel joint Self-Correlation and Cross-Correspondence Estimation Framework (SC$^{3}$EF) consisting of \textcolor{black}{four consecutive stages, including image decomposition, local feature \& self-correlation extraction, local- \& global- correspondence estimation, and hierarchical optical flow estimation.} A convolution-transformer-based pipeline is proposed to gather both local mutual features and global self-correlation cues for effective estimation of dense correspondence maps between unaligned RGB-T multispectral images. Compared with state-of-the-art (SOTA) cross-modality image registration methods \cite{arar2020unsupervised, wang2022unsupervised, zhou2022promoting}, our proposed SC$^{3}$EF can achieve more accurate and robust matching results for RGB-T images with significant misalignment, as illustrated in Fig.~\ref{intro_example}. The main contributions of this paper are summarized as follows:

\begin{enumerate}[(1)]
	

\item \textcolor{black}{We propose a hybrid framework for RGB-T image-dense correspondence estimation and registration. It properly combines the advantages of modeling ability from CNN and Transformer architectures when processing high-frequency (HF) representative and low-frequency (LF) contextual components from unpaired source images.}
	

\item \textcolor{black}{We innovatively design two cross-correspondence estimation modules to provide supplementary information for estimating cross-correspondences on images with significant modal discrepancies based on both local distinct features and global contextual cues, including HF-local and LF-global cross-correspondence estimation modules (LCCE and GCCE).}

\item \textcolor{black}{We also design two frequency-specific self-correlation modeling modules to extract HF-local distinctive features and encode LF-global correlations of intra-modality based on their intrinsic properties, including a convolution-based local feature extraction module (LFE) and a transformer-based global self-correlation extraction module (GSCE).}

\item \textcolor{black}{By simulating how human observers perform the challenging RGB-T image registration task, our proposed SC$^{3}$EF can achieve consistently superior results on four representative RGB-T datasets compared to state-of-the-art methods, demonstrating its effectiveness. Furthermore, it also shows competitive generalization performance on other cross-modal datasets such as RGB-N and RGB-D.}

\end{enumerate}

The rest of this paper is organized as follows: we first review the existing methods for multimodal image registration in Section \ref{related_works}. In Section \ref{approach}, we demonstrate details of our proposed SC$^{3}$EF method including a number of purpose-built convolution-transformer-based modules. Section \ref{exp_res} provides extensive experimental results and analysis to evaluate the performance of SC$^{3}$EF, quantitatively and qualitatively. Finally, Section \ref{conclusion} concludes this paper.

\section{Related works}
\label{related_works}

In this section, we first provide a brief overview of \textcolor{black}{traditional methods \cite{jiang2021review, roche1998correlation, zhao2006image, viola1997alignment, lowe2004distinctive, rublee2011orb, du2018infrared, jiang2020contour, cao2020boosting} for multimodal image registration and then present a review of the SOTA deep learning-based methods \cite{wang2018infrared, dosovitskiy2015flownet, wang2022unsupervised, xu2022rfnet, walters2021there, sun2018pwc, jiang2021review, wang2023msagan, xu2022gmflow, rajalakshmi2023multi, natarajan2022dynamic, song2022cross} for mono- and cross-modal transformation parameter estimation as well as dense correspondence estimation. The typical methods are compared and summarized in Table \ref{related_tab}, highlighting their types, contributions, strengths and limitations.}


Traditional descriptor-based multispectral image registration methods include area- and feature-based ones \cite{jiang2021review}. Given a predefined transformation model along with a similarity metric for the transform parameter optimization, area-based methods utilize the low-level intensity information to perform the image registration. The cross correlation (CC) \cite{roche1998correlation} and normalized correlation coefficient (NCC) \cite{zhao2006image} assumed that there was an intrinsically linear correlation between two source images. One of the most representative metrics, mutual information (MI) \cite{viola1997alignment}, measures mutual dependence and is widely used in cross-modality registration. However, the realistic cross-modality registration cannot satisfy the absolute linearity assumption, which is largely affected by the image quality and the size of the offset area. Feature-based methods (e.g., point, edge, and contour) regard the extracted sparse features as the simplified representation of the whole image. Existing point-based methods, such as scale-invariant feature transform (SIFT) \cite{lowe2004distinctive}, oriented fast and rotated brief (ORB) \cite{rublee2011orb}, and partial intensity invariant feature descriptor (PIIFD) \cite{du2018infrared}, perform poorly on cross-modality registration due to the significant differences between modalities. CAO-C2F  \cite{jiang2020contour} designs a contour angle orientation extraction strategy to increase the orientation similarities of feature points on visible and thermal images of power devices. Furthermore, based on the statistical prior of gradient-intensity correlation in natural images, Cao et al. \cite{cao2020boosting} proposed a structure consistency boosting (SCB) transform to enhance the structural similarity of multispectral images. Nevertheless, the assumption that salient and consistent features exist in multispectral images cannot be easily extended to other complex scenes (e.g., automatic driving and nighttime surveillance), and thus these descriptor-based methods cannot achieve robust and accurate multispectral image registration results. \textcolor{black}{To sum up, traditional cross-modal image registration methods are predominantly based on idealized modality feature priors and hand-crafted feature extraction techniques. These approaches often face challenges in delivering robust and reliable performance, especially in real-world conditions with adverse circumstances.}

\begin{table}[tb]
	\Large
	\centering
	\renewcommand{\arraystretch}{1.2}
	\caption{\textcolor{black}{Illustration of the data flow and dimensional configurations at each major stage: image decomposition, local feature \& self-correlation extraction, local- \& global- correspondence estimation and hierarchical optical flow estimation.}}
	\resizebox{0.48 \textwidth}{!}{
		\begin{tabular}{cccccc}
			\hline
			\multicolumn{1}{c}{No.} &
			\multicolumn{1}{c}{Module} &
			\multicolumn{1}{c}{Input} &
			\multicolumn{1}{c}{Input Size} &
			\multicolumn{1}{c}{Output} &
			\multicolumn{1}{c}{Output Size} \\
			\hline\hline
			\multicolumn{1}{c}{1} & \multicolumn{5}{c}{Image Decomposition} \\
			\hline
			\rule{0pt}{4ex} ~ & $\boldsymbol{\mathcal{M}}_{GF} \times 2$ & $I_{v}, I_{t}$ & $(3, H, W)$ & \makecell{$I_{v}^{hf}, I_{v}^{lf}$, \\ $I_{t}^{hf}, I_{t}^{lf}$} & $(3, H, W)$ \\
			\hline\hline
			\multicolumn{1}{c}{2} & \multicolumn{5}{c}{Local Feature \& Self-Correlation Extraction} \\
			\hline
			\rule{0pt}{4ex} ~ & $\boldsymbol{\mathcal{M}}_{LFE} \times 2$ & $I_{v}^{hf}, I_{t}^{hf}$ & $(3, H, W)$ & $F_{v}, F_{t}$ & $(96, \dfrac{H}{8}, \dfrac{W}{8})$ \\
			\rule{0pt}{4ex} ~ & $\boldsymbol{\mathcal{M}}_{GSCE} \times 2$ & $I_{v}^{lf}, I_{t}^{lf}$ & $(3, H, W)$ & $\mit\Phi_{v}, \mit\Phi_{t}$ & $(96, \dfrac{H}{8}, \dfrac{W}{8})$ \\
			\hline\hline
			\multicolumn{1}{c}{3} & \multicolumn{5}{c}{Local- \& Global- Correspondence Estimation} \\
			\hline
			\rule{0pt}{4ex} ~ & $\boldsymbol{\mathcal{M}}_{LCCE} \times 2$ & \makecell{$[F_{v} || F_{t}]$, \\ $[F_{t} || F_{v}]$} & $(96, \dfrac{H}{8}, \dfrac{W}{8})$ & $F_{v-t}, F_{t-v}$ & $(384, \dfrac{H}{32}, \dfrac{W}{32})$ \\
			\rule{0pt}{4ex} ~ & $\boldsymbol{\mathcal{M}}_{GCCE} \times 2$ & \makecell{$[\mit\Phi_{v} || F_{v}]$, \\ $[\mit\Phi_{t} || F_{t}]$} & $(96, \dfrac{H}{8}, \dfrac{W}{8})$ & $\mit\Phi_{v-t}, \mit\Phi_{t-v}$ & $(384, \dfrac{H}{32}, \dfrac{W}{32})$ \\
			\hline\hline
			\multicolumn{1}{c}{4} & \multicolumn{5}{c}{Hierarchical Optical Flow Estimation} \\
			\hline
			\rule{0pt}{4ex} ~ & $\boldsymbol{\mathcal{M}}_{FRB} \times 5$ & \makecell{$[F_{v-t} || F_{t-v}]$, \\ $[\mit\Phi_{v-t} || \mit\Phi_{t-v}]$} & $(384, \dfrac{H}{32}, \dfrac{W}{32})$ & $I_{f}$ & $(2, H, W)$ \\
			\hline
	\end{tabular}}
	\label{data_flow}
\end{table}

\textcolor{black}{In contrast, deep learning-based dense correspondence estimation approaches have been increasingly adopted for cross-modal registration tasks due to their superior capabilities for feature extraction and representation.} To bridge the gap between visible and thermal images, Wang et al. \cite{wang2018infrared} proposed a two-stage adversarial network that contained a style transfer sub-network to convert the visible image to the thermal domain and a FlowNet-based \cite{dosovitskiy2015flownet} transform estimation to warp the thermal image, respectively. Similarly, Wang et al. \cite{wang2022unsupervised} proposed the cross-modality generation-registration paradigm (CGRP), utilizing a style transfer network first and incorporating the mapped pseudo-thermal image into the flowing registration network to estimate the dense flow field. Furthermore, Xu et al. \cite{xu2022rfnet} proposed a joint multi-modality image registration and fusion framework to promote registration accuracy through the fusion results. Walter et al. \cite{walters2021there} proposed a cross-spectral flow field estimation framework, that incorporated a modified PWC-Net \cite{sun2018pwc} as the main flow estimation module into a cycle-GAN-based architecture to address the multispectral registration problem. Moreover, they utilized a perceptual loss to constrain the structural consistency across the modalities. The generated pseudo thermal image by the style transfer network differs significantly from the real-captured ones when there is no prior knowledge of thermal radiation in specific scenes. As a result, converting the challenging multispectral image registration problem into a mono-modality image alignment task becomes difficult \cite{jiang2021review}. Moreover, convolutional backbones, which are commonly used, only extract local representative features in predefined windows, making it difficult to address large displacements in unaligned visible and thermal images \cite{xu2022gmflow}.

\begin{figure}[tb]
	\normalsize
	\centering
	\includegraphics[scale=0.36]{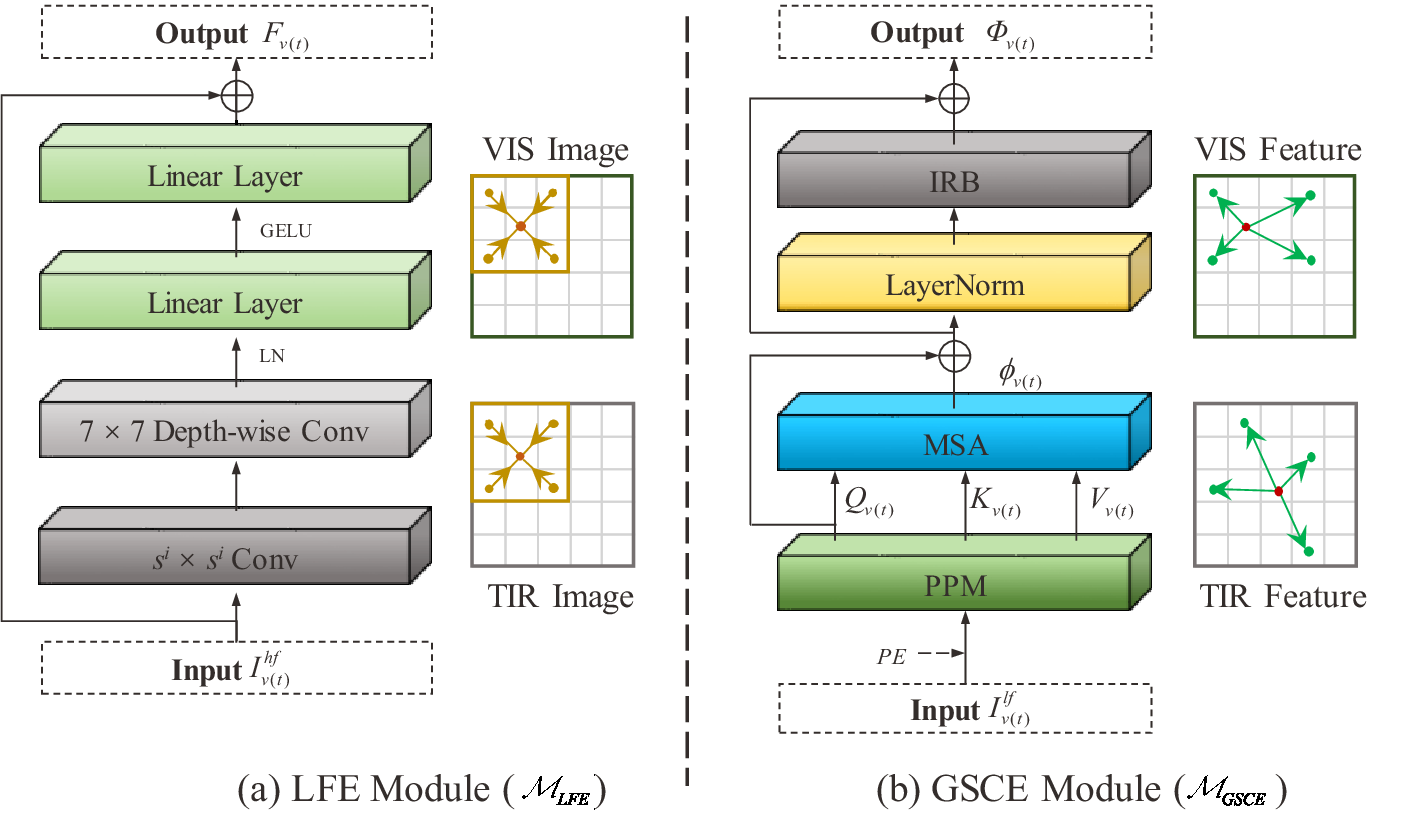}
	\caption{\textcolor{black}{Detailed structures for the proposed (a) LFE and (b) GSCE modules.}}
	\label{LFE_GSCE_fig}
\end{figure}

\section{Our approach}
\label{approach}

In this paper, we present SC$^{3}$EF to perform dense correspondence estimation between visible and thermal images. The overall framework of the proposed convolution-transformer-based SC$\rm ^{3}$EF model is illustrated in Fig.~\ref{overall_arch}. \textcolor{black}{Given the unaligned visible $I_{v}$ and thermal image $I_{t}$ pairs as input, a standard edge-preserving filter $\mathcal{M}_{GF}$ (e.g., the guided filter \cite{he2012guided})} is deployed to extract the low-frequency (LF) components ($I^{lf}_{v}$, $I^{lf}_{t}$) and high-frequency (HF) components ($I^{hf}_{v}$, $I^{hf}_{t}$). Then, the extracted LF and HF components ($I^{hf}_{v}$ and $I^{lf}_{v}$, $I^{hf}_{t}$ and $I^{lf}_{t}$) are fed to the \textbf{self-correlation extraction stage}, consisting of several stacked convolution-based Local Feature Extraction (LFE) and transformer-based Global Self-Correlation Extraction (GSCE) modules (details are provided in Section \ref{LFE-GSCE-Section}), to extract local distinctive features and encode global correlations of intra-modality, respectively. Here, the distinct features extracted in HF components are incorporated into the corresponding LF streams as complementary information to achieve a robust estimation of global self-correlation. In the consecutive \textbf{cross-correspondence estimation stage}, a number of Local Cross-Correspondence Estimation (LCCE) and Global Cross-Correspondence Estimation (GCCE) modules (details are provided in Section \ref{LCCE-GCCE-Section}) are stacked to estimate correspondences between unaligned visible and thermal images based on both local representative features and global contextual cues. The correspondences based on the matching of local features are generated along the channel dimension of concatenated visible and thermal images through the LCCE modules. Moreover, the encoded visible and thermal self-correlations are fused and fed to the stacked GCCE modules to search for correspondences across modalities based on important self-correlation cues. The local correspondences generated by LCCE modules are infused into the corresponding GCCE modules as auxiliary information, enabling interactive refinement between local and global correspondence estimation. In the \textbf{hierarchical optical flow estimation stage}, a differentiable matching layer \cite{xu2020aanet} is deployed to estimate the initial optical flow map based on the computed cross-spectral correspondences (details are provided in Section \ref{HFE}). The predicted optical flow maps based on both local representative features and global contextual cues are merged through weighted element-wise addition, and then progressively refined from 1/32 of the original input size to full size using several stacked flow refinement blocks (FRB) (details are provided in Section \ref{HFE}). \textcolor{black}{The detailed data flow and dimensional configurations of the major stages are depicted in Table~\ref{data_flow}.}

\begin{figure}[tb]
	\normalsize
	\centering
	\includegraphics[scale=0.38]{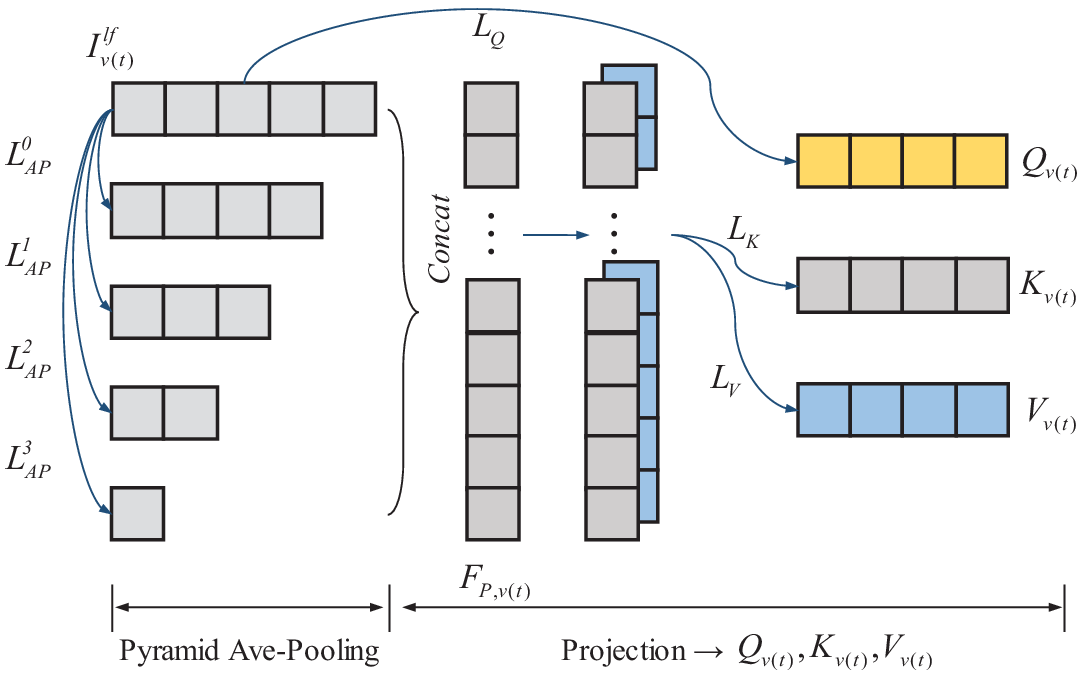}
	\caption{\textcolor{black}{Details of the proposed PPM that mainly consists of two stages: pyramid average-pooling and linear projection.}}
	\label{module_PPM}
\end{figure}

\subsection{Local Feature and Global Self-Correlation Extraction}
\label{LFE-GSCE-Section}

By simulating human observers, we present convolution-based LFE and transformer-based GSCE modules, as shown in Fig.~\ref{LFE_GSCE_fig}. We introduce the depth-wise convolution (DWConv) block to separately capture the local distinct and mutual feature representations from HF visible and thermal images in the LFE module. Besides, given the success of the transformer-based architectures \cite{vaswani2017attention} in various downstream computer vision tasks, we employ the qkv self-attention to extract the global self-correlation in the proposed GSCE module.

In the LFE module ($\boldsymbol{\mathcal{M}}_{LFE}$), we aim to extract representative features from HF streams by utilizing overlapped windows in the sliding operation of convolutions. As shown in Fig.~\ref{LFE_GSCE_fig} (a), given the HF component ($I_{v(t)}^{hf}$) as input of \textcolor{black}{$\boldsymbol{\mathcal{M}}_{LFE}$, we firstly utilize a $s \times s$ vanilla convolution ($s=4, 8$)} to downsize the input image to $1/4$ and $1/8$, matching the shape of the corresponding feature map generated by GSCE modules. Then, we deploy a $7 \times 7$ DWConv to separately extract the local HF features within each channel. Considering the detrimental effect during training the model \cite{wu2021rethinking}, we replace the typical BatchNorm (BN) with Layer Normalization (LN) \cite{ba2016layer}. Next, we deploy two linear layers to execute the feature aggregation, which is equivalent to $1 \times 1$ convolution. Besides replicating the style of a transformer block, we utilize the Gaussian Error Linear Unit (GELU) \cite{hendrycks2016gaussian}, which can be regarded as a smoother variant of ReLU. Finally, a shortcut connection is introduced to facilitate the convergence of training. The extracted HF features (\textcolor{black}{$F_{v(t)}$}) can be formulated as

\begin{figure}[tb]
	\normalsize
	\centering
	\includegraphics[scale=0.36]{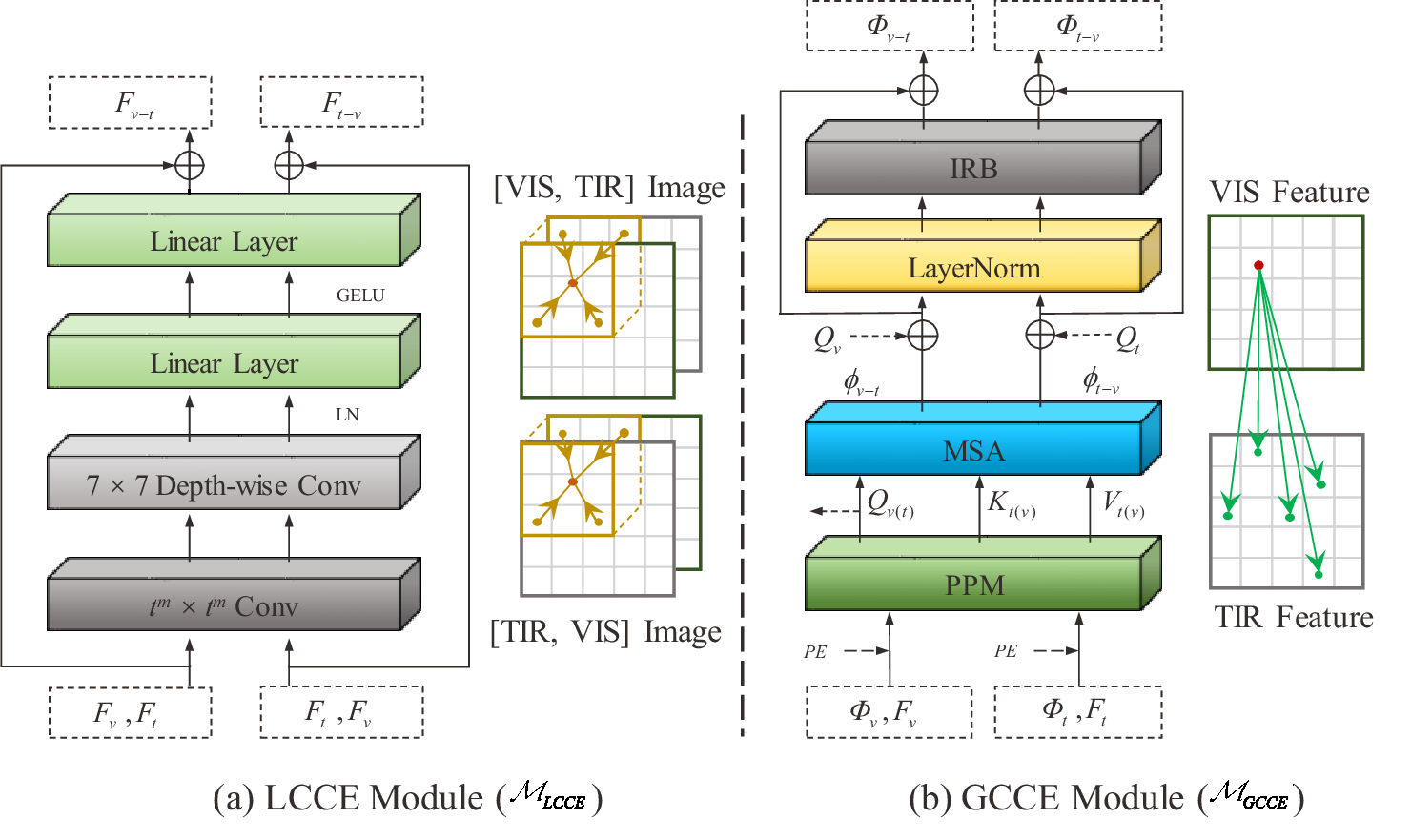}
	\caption{\textcolor{black}{Detailed structures for the proposed (a) LCCE and (b) GCCE modules.}}
	\label{GCCE-LF-LFEB-HF-Fig}
\end{figure}

\begin{eqnarray}
\textcolor{black}{F_{v(t)} = \boldsymbol{\mathcal{M}}_{LFE}(I^{hf}_{v(t)}),}
\end{eqnarray}

\noindent where \textcolor{black}{$\boldsymbol{\mathcal{M}}_{LFE}$ denotes the LFE module.} It is worth mentioning that the extracted \textcolor{black}{$F_{v(t)}$} is infused with the input of the GSCE module to further facilitate global self-correlation extraction.

In the GSCE module ($\boldsymbol{\mathcal{M}}_{GSCE}$), we leverage the self-attention mechanism of transformer \cite{vaswani2017attention} to establish the global self-correlation of intra-modality, which is the important guidance to seek the alignment across modality in the following cross-correspondence stage, and the full process of global self-correlation extraction is formulated as

\begin{equation}
	\textcolor{black}{
	\mit\Phi_{v(t)} = \boldsymbol{\mathcal{M}}_{\text{IRB}}\left( \boldsymbol{\mathcal{L}}\boldsymbol{\mathcal{N}}\left( \boldsymbol{\mathcal{M}}_{\text{MSA}}\left( \boldsymbol{\mathcal{M}}_{\text{PPM}}\left( I^{lf}_{v(t)} \right) \right) \right) \right)}
\end{equation}

\noindent \textcolor{black}{where $\boldsymbol{\mathcal{M}}_{\text{PPM}}$, $\boldsymbol{\mathcal{M}}_{\text{MSA}}$ and $\boldsymbol{\mathcal{M}}_{\text{IRB}}$ denote our proposed pyramid pooling module (PPM), multi-head self-attention (MSA) \cite{dosovitskiy2020image} and the inverted bottleneck block \cite{sandler2018mobilenetv2}, respectively. $I^{lf}_{v(t)}$ and $\mit\Phi_{v(t)}$ denote the input of GSCE module and the final output feature, respectively.}


As illustrated in Fig.~\ref{LFE_GSCE_fig} (b), we first split the input image \textcolor{black}{$I^{lf}_{v(t)}$} into patches to perform patch embedding (PE). In order to ensure the feature continuity across adjacent patches, we adopt a vanilla convolution to extract the overlapped feature patches and embed them in a patch sequence, referring to \cite{wang2022pvt}. Given the patch embedding, we aim to generate multi-scale contextual prior with condensed representation to improve the self-attention ability on the basis of reducing the computational load of the transformer. To this end, \textcolor{black}{we introduce a pyramid pooling module (PPM) embedded within the transformer layer to generate the multi-scale feature maps by applying multiple average pooling layers with different ratios onto the low-frequency image components of visible and thermal images, as illustrated in Fig.~\ref{module_PPM}. It effectively reduces the computational burden while simultaneously capturing representative contextual information.} Different from the traditional qkv acquisition strategy, we utilized the concatenated pyramid features to generate \textcolor{black}{query $Q_{v(t)}$, key $K_{v(t)}$ and value $V_{v(t)}$} by

\begin{equation}
\textcolor{black}{Q_{v(t)},K_{v(t)},V_{v(t)} =  \boldsymbol{L}_{Q}(I^{lf}_{v(t)}),\boldsymbol{L}_{K}(F_{P,v(t)}),\boldsymbol{L}_{V}(F_{P,v(t)})}
\end{equation}

\noindent where $\boldsymbol{L}_{Q}$, $\boldsymbol{L}_{K}$ and $\boldsymbol{L}_{V}$ denote the different weighted linear transformation layer, and \textcolor{black}{$F_{P,v(t)}$ denote the feature after pyramid average pooling and the extracted condensed feature, which is the output of PPM.} Subsequently, to capture global attention, the MSA in the GSCE module is defined as

\begin{eqnarray}
\textcolor{black}{\mit\phi_{v(t)} = \operatorname{Softmax}(\frac{Q_{v(t)}(K_{v(t)})^{T}}{\sqrt{C_{K_{v(t)}}}}) \times V_{v(t)}}
\end{eqnarray}

\noindent where \textcolor{black}{$C_{K_{v(t)}}$} and \textcolor{black}{$\mit\phi_{v(t)}$} denote the dimension of \textcolor{black}{$K_{v(t)}$} and the computed self-attention map, respectively.

%

\begin{figure}[tb]
	\normalsize
	\centering
	\includegraphics[scale=0.67]{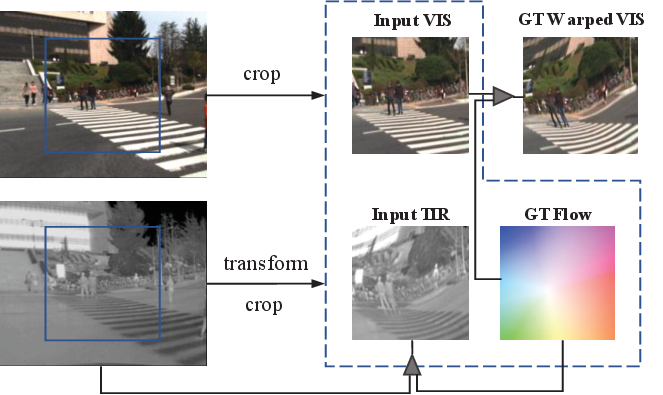}
	\caption{Illustration of the simulated unaligned data generation procedure with samples.}
	\label{data_gen}
\end{figure}

\subsection{Local and Global Cross-Correspondence Estimation}
\label{LCCE-GCCE-Section}

Given the encoded global self-correlation of each modality, our objective is to estimate the cross-correspondences of inter-modality features from HF and LF streams in parallel. As shown in Fig.~\ref{GCCE-LF-LFEB-HF-Fig}, we extend LFE module to the cross-spectral domain as LCCE module and propose a novel modified transformer-based GCCE module to jointly perform the cross-correspondence estimation for distinct features and global image content.

In the LCCE module ($\boldsymbol{\mathcal{M}}_{LCCE}$), we aim to learn the cross-correspondence of distinct features between HF-visible and HF-thermal channels to provide the supplementary information for the corresponding LF stage and estimate the optical flow map of HF stream. Concretely, to capture the cross-spectral correspondence, we directly concatenate \textcolor{black}{HF visible and thermal features as inputs, and then we utilize a $t \times t$ vanilla convolution ($t=16, 32$)} to downsize the input image to $1/16$ and $1/32$, matching the shape of corresponding feature map generated from the GCCE module. The following process is the same as the LFE module, which can be formulated as

\begin{eqnarray}
\textcolor{black}{F_{v-t(t-v)} = \boldsymbol{\mathcal{M}}_{LCCE}([F_{v}||F_{t})], [F_{t}||F_{v}])}
\end{eqnarray}

\noindent where \textcolor{black}{$\boldsymbol{\mathcal{M}}_{LCCE}$} denotes the LCCE module, while \textcolor{black}{$F_{v-t}$ and $F_{t-v}$} represent the computed cross-correspondences of visible and thermal features, respectively, by the LCCE module. Furthermore, the estimated \textcolor{black}{$F_{v-t(t-v)}$} are infused with the inputs of the GCCE module to further facilitate global cross-correspondence estimation.

In the GCCE module ($\boldsymbol{\mathcal{M}}_{GCCE}$), we carefully modify the self-correlation extraction mechanism proposed in the GSCE module to adapt the cross-spectral correspondence estimation. The full process of global cross-correspondence estimation can be formulated as

\begin{equation}
\textcolor{black}{
\mit\Phi_{v-t(t-v)} = \boldsymbol{\mathcal{M}}_{GCCE}([\Phi_{v} || F_{v}], [\Phi_{t} || F_{t}])}
\end{equation}

\noindent \textcolor{black}{where $\mit\Phi_{v}$, $\mit\Phi_{t}$ and $\mit\Phi_{v-t(t-v)}$ denote the input visible, thermal and final output features of the GCCE module, respectively.}



Specifically, \textcolor{black}{query $Q_{v(t)}$, key $K_{t(v)}$ and value $V_{t(v)}$ are also generated from PPM, where $Q_{v(t)}$, $K_{t(v)}$ and $V_{t(v)}$ are derived from different channels (e.g, $Q_{v(t)}$ from the visible channel, $K_{t(v)}$ and $V_{t(v)}$ from the thermal channel), to realize the cross-correspondence estimation.} Then, the cross-spectral MSA can be formulated as

\begin{eqnarray}
\textcolor{black}{\mit\phi_{v-t(t-v)} = \operatorname{Softmax}(\frac{Q_{v(t)}(K_{t(v)})^{T}}{\sqrt{C_{K_{t(v)}}}}) \times V_{t(v)}}
\end{eqnarray}

\noindent where \textcolor{black}{$\mit\phi_{v-t(t-v)}$ denotes the computed cross-attention map.}

\begin{table}[tb]
	\Large
	\renewcommand{\arraystretch}{1.2}
	\centering
	\caption{\textcolor{black}{Detailed parameter settings of the main modules. Here, $D_{in}$, $D_{out}$, $K$, $S$, and $P$ represent the input and output feature dimensions, convolution kernel size, stride, and padding, respectively. The numbers in "[]" correspond one-to-one with the module quantities, where $P.R.$ denotes the pyramid pooling ratio.}}
	\resizebox{0.49 \textwidth}{!}{
		\begin{tabular}{ccccccc}
			\hline
			\multicolumn{1}{c}{Module} & 
			\multicolumn{1}{c}{Sub-Module} &
			\multicolumn{1}{c}{$D_{in}$} &
			\multicolumn{1}{c}{$K$} &
			\multicolumn{1}{c}{$S$} &
			\multicolumn{1}{c}{$P$} &
			\multicolumn{1}{c}{$D_{out}$} \\
			\hline
			\multirow{2}{*}{$\boldsymbol{\mathcal{M}}_{LFE} \times 2$} & \multicolumn{1}{c}{\textit{Basic Conv.}} & $[3, 48]$ & $[4, 8]$ & $[4, 8]$ & $[0, 0]$ & $[48, 96]$ \\
			~ & \multicolumn{1}{c}{\textit{DWConv}} & $[48, 96]$ & $[7, 7]$ & $[1, 1]$ & $[3, 3]$ & $[48, 96]$ \\
			\multirow{2}{*}{$\boldsymbol{\mathcal{M}}_{GSCE} \times 2$} & \multicolumn{1}{c}{\textit{PPM}} & $[48, 96]$ & \multicolumn{3}{c}{P.R.: $([12, 16, 20, 24], [6, 8, 10, 12])$} & $[48, 96]$\\
			~ & \multicolumn{1}{c}{\textit{IRB}} & $[48, 96]$ & $[7, 7]$ & $[1, 1]$ & $[3, 3]$ & $[48, 96]$ \\
			\multirow{2}{*}{$\boldsymbol{\mathcal{M}}_{LCCE} \times 2$}  & \multicolumn{1}{c}{\textit{Basic Conv.}} & $[3, 48]$ & $[4, 8]$ & $[4, 8]$ & $[0, 0]$ & $[48, 96]$ \\
			~ & \multicolumn{1}{c}{\textit{DWConv}} & $[48, 96]$ & $[7, 7]$ & $[1, 1]$ & $[3, 3]$ & $[48, 96]$ \\
			\multirow{2}{*}{$\boldsymbol{\mathcal{M}}_{GCCE} \times 2$} & \multicolumn{1}{c}{\textit{PPM}} & $[48, 96]$ & \multicolumn{3}{c}{P.R.: $([3, 4, 5, 6], [1, 2, 3, 4])$} & $[48, 96]$\\
			~ & \multicolumn{1}{c}{\textit{IRB}} & $[48, 96]$ & $[7, 7]$ & $[1, 1]$ & $[3, 3]$ & $[48, 96]$ \\
			\multirow{4}{*}{$\boldsymbol{\mathcal{M}}_{FRB} \times 5$} & \multicolumn{1}{c}{\textit{Up-sample}} & $[2] \times 5$ & \multicolumn{3}{c}{$[2] \times 5$} & $[2] \times 5$ \\
			~ & \textit{Basic Conv.} & $[2] \times 5$ & $[3] \times 5$ & $[1] \times 5$ & $[1] \times 5$ & $[64] \times 5$ \\
			~ & \textit{Basic Conv.} & $[2] \times 5$ & $[3] \times 5$ & $[1] \times 5$ & $[1] \times 5$ & $[64] \times 5$ \\
			~ & \textit{Basic Conv.} & $[2] \times 5$ & $[3] \times 5$ & $[1] \times 5$ & $[1] \times 5$ & $[64] \times 5$ \\
			\hline
	\end{tabular}}
	\label{sub_module_params}
\end{table}

\begin{table*}[tb]
	\small
	\renewcommand{\arraystretch}{1.2}
	\centering
	\caption{Quantitative comparison to the SOTA methods on the KAIST dataset \cite{hwang2015multispectral}. The best and second-best results are marked in \textbf{bold} and \underline{underlined} respectively. ``$\mathbf{\uparrow}$'' denotes the higher the better, ``$\mathbf{\downarrow}$'' denotes the lower the better.}
	\resizebox{0.91\textwidth}{!}{
		\begin{tabular}{cccccccccc}
			\hline
			\multicolumn{2}{c}{Metrics} &
			\multicolumn{1}{c}{NTG \cite{chen2017normalized}} & 
			\multicolumn{1}{c}{SCB \cite{cao2020boosting}} &
			\multicolumn{1}{c}{DGC-Net \cite{melekhov2019dgc}} &
			\multicolumn{1}{c}{NeMAR \cite{arar2020unsupervised}} &
			\multicolumn{1}{c}{GMFlow \cite{xu2022gmflow}} &
			\multicolumn{1}{c}{CMF \cite{zhou2022promoting}} &
			\multicolumn{1}{c}{UMF-CMGR \cite{wang2022unsupervised}}&
			\multicolumn{1}{c}{SC$^{3}$EF (ours)} \\
			\hline
			\multicolumn{2}{c}{AEPE $\mathbf{\downarrow}$} & 98.87 & 108.53 & 14.65 & 32.69 & \underline{11.85} & 16.30 & 32.45 & \textbf{4.65} \\
			\multirow{3}{*}{PCK ($\%$) $\mathbf{\uparrow}$} & \multicolumn{1}{c}{$1 px$} & 2.23 & 1.74 & 8.16 & 2.84 & 9.72 & \underline{12.85} & 3.28 & \textbf{23.6} \\
			~ & $3 px$ & 6.48 & 5.37 & 23.89 & 8.48 & 28.23 & \underline{33.27} & 9.90 & \textbf{62.11} \\
			~ & $5 px$ & 10.48 & 8.85 & 37.90 & 14.06 & 44.34 & \underline{47.09} & 16.34 & \textbf{83.61} \\
			\multicolumn{2}{c}{CC $\mathbf{\uparrow}$} & 0.59 & 0.70 & 0.77 & 0.68 & \underline{0.81} & 0.79 & 0.70 & \textbf{0.88} \\
			\multicolumn{2}{c}{NCC $\mathbf{\uparrow}$} & 0.11 & 0.14 & 0.16 & 0.12 & \underline{0.18} & \underline{0.18} & 0.13 & \textbf{0.24} \\
			\multicolumn{2}{c}{MI $\mathbf{\uparrow}$} & 6.12 & 5.72 & \underline{7.03} & 6.95 & 6.55 & 6.57 & \textbf{7.10} & 6.64 \\
			\multicolumn{2}{c}{PSNR $\mathbf{\uparrow}$} & 10.63 & 11.55 & \underline{14.41} & 12.93 & 14.33 & 13.19 & 13.10 & \textbf{16.94} \\
			\multicolumn{2}{c}{SCD $\mathbf{\uparrow}$} & 0.15 & 0.34 & 0.50 & 0.29 & \underline{0.56} & 0.48 & 0.35 & \textbf{0.72} \\
			\multicolumn{2}{c}{SSIM $\mathbf{\uparrow}$} & 0.41 & 0.44 & 0.53 & 0.51 & \underline{0.59} & 0.55 & 0.50 & \textbf{0.68} \\
			\hline
	\end{tabular}}
	\label{kaist_com}
\end{table*}

\begin{figure*}[tb]
	\normalsize
	\centering
	\includegraphics[scale=0.41]{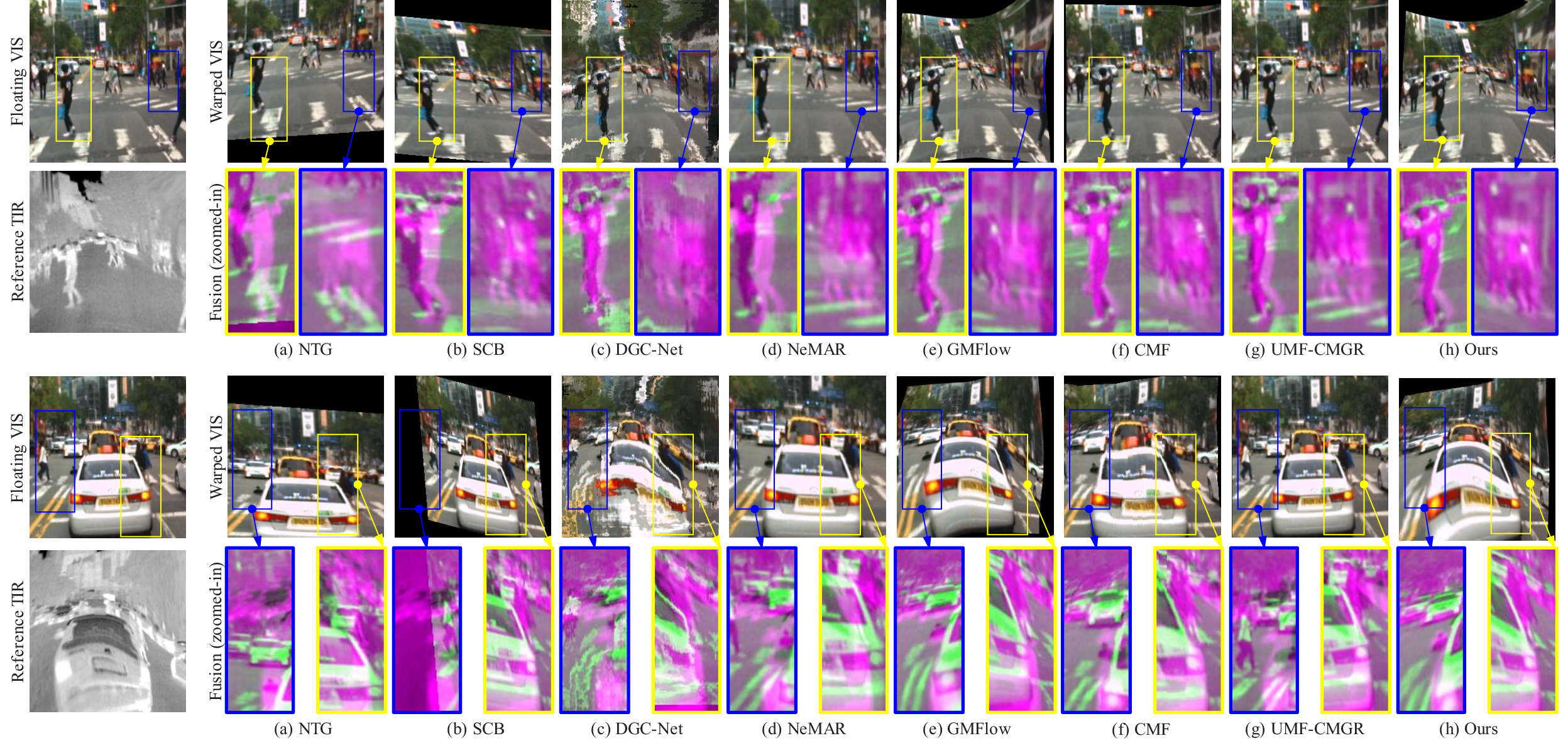}
	\caption{\textcolor{black}{Visual comparison of SOTA methods and SC$^{3}$EF on the KAIST dataset \cite{hwang2015multispectral}. The warped visible images, paired with their corresponding zoomed-in fusion results highlighted in yellow and blue boxes, are presented to visualize the misalignments.}}
	\label{kaist_res}
\end{figure*}

\subsection{Hierarchical Optical Flow Estimation}
\label{HFE}

As illustrated in Fig.~\ref{overall_arch}, to generate the optical flow map from estimated cross-correspondences of HF and LF streams, we utilize the differentiable \textcolor{black}{matching layer ($\boldsymbol{ML})$} \cite{kendall2017end} to compute the pixel-level matching distribution and generate the optical flow map reflecting the corresponding pixel coordinate difference. Then, the \textcolor{black}{final optical flow map ($I_{f}$) can be computed by five stacked $\boldsymbol{\mathcal{M}}_{FRB}$ modules, which can be formulated as}

\begin{eqnarray}
\textcolor{black}{I_{f} = \mathcal{M}_{FRB}( \boldsymbol{ML}(F_{v-t},F_{t-v})+\boldsymbol{ML}(\mit\Phi_{v-t}, \mit\Phi_{t-v}))}
\end{eqnarray}

\noindent \textcolor{black}{where each $\boldsymbol{\mathcal{M}}_{FRB}$ module is composed of an up-sample layer connected with three basic convolutional layers in series.}




\subsection{Loss Function} 
Given an image pair ($I_{v}$, $I_{t}$) and the ground truth (GT) optical flow maps $I_{f}^{gt, k}$ \textcolor{black}{generated by each $\boldsymbol{\mathcal{M}}_{FRB}$ with different scales}, we define an objective loss function to supervise the hierarchical optical flow estimations using the EPE loss \cite{dosovitskiy2015flownet} at every pyramid level as

\begin{eqnarray}
\textcolor{black}{L = \sum_{k=0}^{N} \alpha^{N-k}\left\| I^{k}_{f} - I_{f}^{gt, k} \right\|_{1}}
\end{eqnarray}

\noindent where $N$ denotes the number of estimated optical flow maps ($I^{k}_{f}$) by hierarchical optical flow estimation stage (here $N=5$), and $\alpha$ is an exponentially increasing weight to give higher weights for the larger size of predictions which is set to 0.9.

\begin{table*}[tb]
	\small
	\renewcommand{\arraystretch}{1.2}
	\centering
	\caption{Quantitative comparison to the SOTA methods on the RoadScene dataset \cite{xu2020u2fusion}. The best and second-best results are marked in \textbf{bold} and \underline{underlined} respectively. ``$\mathbf{\uparrow}$'' denotes the higher the better, ``$\mathbf{\downarrow}$'' denotes the lower the better.}
	\resizebox{0.91\textwidth}{!}{
		\begin{tabular}{cccccccccc}
			\hline
			\multicolumn{2}{c}{Metrics} &
			\multicolumn{1}{c}{NTG \cite{chen2017normalized}} & 
			\multicolumn{1}{c}{SCB \cite{cao2020boosting}} &
			\multicolumn{1}{c}{DGC-Net \cite{melekhov2019dgc}} &
			\multicolumn{1}{c}{NeMAR \cite{arar2020unsupervised}} &
			\multicolumn{1}{c}{GMFlow \cite{xu2022gmflow}} &
			\multicolumn{1}{c}{CMF \cite{zhou2022promoting}} &
			\multicolumn{1}{c}{UMF-CMGR \cite{wang2022unsupervised}}&
			\multicolumn{1}{c}{SC$^{3}$EF (ours)} \\
			\hline
			\multicolumn{2}{c}{AEPE $\mathbf{\downarrow}$} & 88.45 & 56.58 & 31.21 & 32.14 & 21.49 & \underline{17.56} & 34.40 & \textbf{9.29} \\
			\multirow{3}{*}{PCK ($\%$) $\mathbf{\uparrow}$} & \multicolumn{1}{c}{$1 px$} & 2.47 & 2.16 & 3.09 & 3.11 & 4.72 & \textbf{14.46} & 3.08 & \underline{13.90} \\
			~ & $3 px$ & 7.16 & 6.55 & 9.23 & 9.30 & 14.07 & \underline{34.58} & 9.21 & \textbf{38.74} \\
			~ & $5 px$ & 11.53 & 10.76 & 15.31 & 15.35 & 23.17 & \underline{47.23} & 15.21 & \textbf{57.88} \\
			\multicolumn{2}{c}{CC $\mathbf{\uparrow}$} & 0.65 & 0.77 & 0.71 & 0.71 & 0.79 & \underline{0.81} & 0.71 & \textbf{0.87} \\
			\multicolumn{2}{c}{NCC $\mathbf{\uparrow}$} & 0.09 & 0.14 & 0.12 & 0.12 & 0.14 & \underline{0.18} & 0.12 & \textbf{0.20} \\
			\multicolumn{2}{c}{MI $\mathbf{\uparrow}$} & 5.52 & 5.88 & 6.47 & \underline{6.54} & 6.10 & 6.10 & \textbf{6.57} & 6.08 \\
			\multicolumn{2}{c}{PSNR $\mathbf{\uparrow}$} & 10.72 & 12.28 & \underline{14.66} & 14.60 & 13.67 & 13.39 & 14.54 & \textbf{15.86} \\
			\multicolumn{2}{c}{SCD $\mathbf{\uparrow}$} & 0.21 & 0.38 & 0.38 & 0.38 & 0.47 & \underline{0.49} & 0.38 & \textbf{0.64} \\
			\multicolumn{2}{c}{SSIM $\mathbf{\uparrow}$} & 0.43 & 0.51 & 0.53 & 0.52 & 0.54 & \underline{0.58} & 0.52 & \textbf{0.70} \\
			\hline
	\end{tabular}}
	\label{road_com}
\end{table*}

\begin{figure*}[tb]
	\normalsize
	\centering
	\includegraphics[scale=0.41]{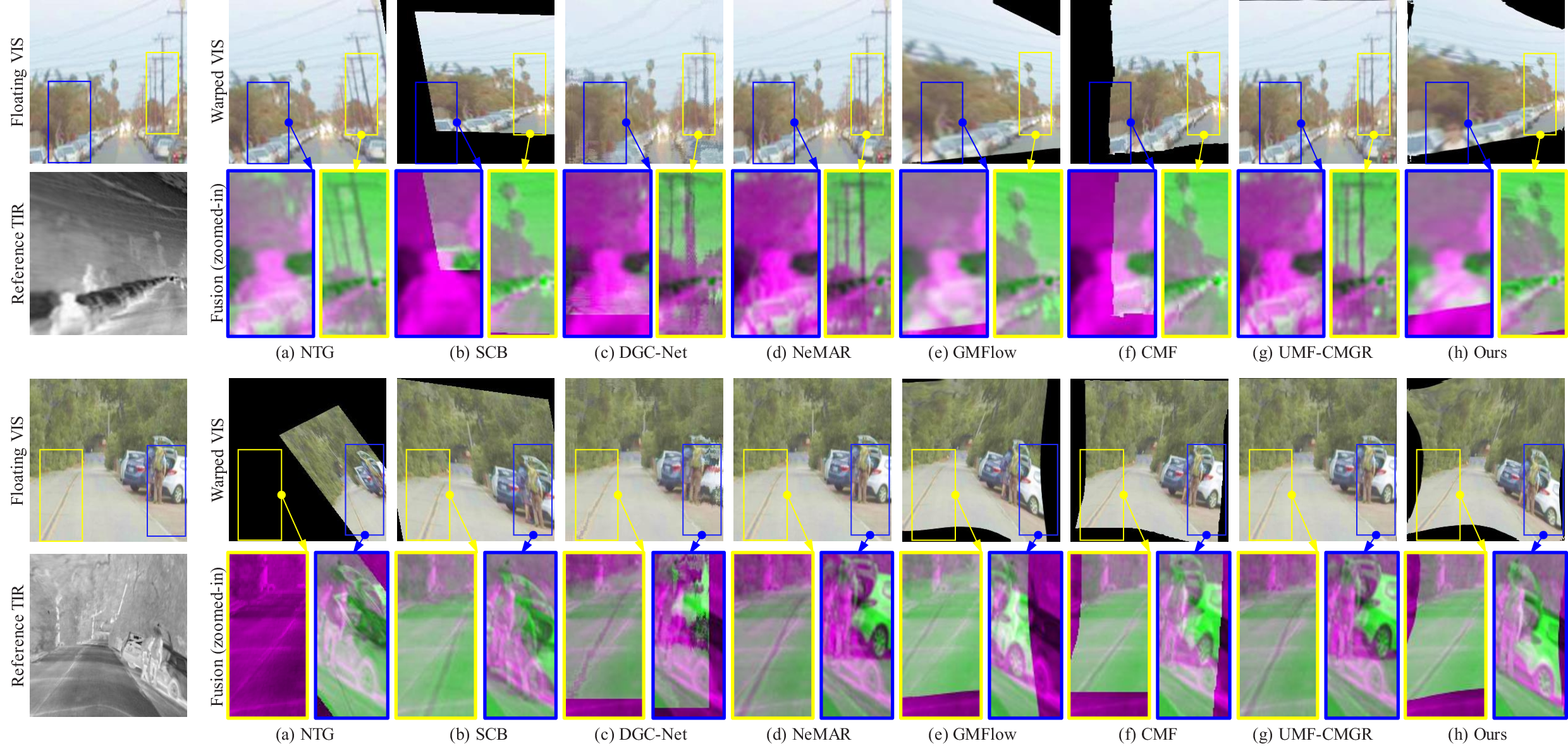}
	\caption{\textcolor{black}{Visual comparison of SOTA methods and SC$^{3}$EF on the RoadScene dataset \cite{xu2020u2fusion}. The warped visible images, paired with their corresponding zoomed-in fusion results highlighted in yellow and blue boxes, are presented to visualize the misalignments.}}
	\label{roadscene_res}
\end{figure*}

\section{Experiments}
\label{exp_res}

\subsection{Experimental settings}
\label{exp_setting}

\textbf{Datasets:} Our comparison experiments are conducted on four well-aligned RGB-T datasets including KAIST \cite{hwang2015multispectral}, RoadScene \cite{xu2020u2fusion}, \textcolor{black}{MFNet \cite{ha2017mfnet}, and M3FD \cite{liu2022target}}. As illustrated in Fig.~\ref{data_gen}, we first randomly generated a number of affine (Aff), homography (HG) and thin-plate spline (TPS) transformation matrices and the corresponding GT optical flow maps, following the data synthetic strategy proposed by \cite{melekhov2019dgc}. Then we obtained a number of warped thermal images by applying the simulated transformation matrices. For the KAIST dataset, we performed data cleansing to remove degraded images such as nighttime visible images and noisy/blurry thermal images. In total, we utilized 17694 image triplets (simulated unaligned visible/thermal images and the corresponding GT optical flow maps) to obtain 16694 triplets as the training dataset and 1000 triplets as the testing dataset. For the RoadScene dataset (221 pairs of visible and thermal images), we generated 221 synthetic triplets and split them into 200 triplets as the training dataset and 21 triplets as the testing dataset. \textcolor{black}{For the MFNet dataset (855 pairs), we generated 855 synthetic triplets and split them into 650 triplets as the training dataset and 205 triplets as the testing dataset. For the M3FD dataset (300 pairs), we generated 300 synthetic triplets and split them into 225 triplets as the training dataset and 75 triplets as the testing dataset.} The simulated dataset, including unaligned RGB-T images and the corresponding GT optical flow maps.


\textbf{Implementation Details:} \textcolor{black}{The detailed parameter settings for our proposed modules and corresponding sub-modules are listed in Table~\ref{sub_module_params}. It includes the input and output feature dimensions at each stage, as well as the primary parameter settings for each sub-module.} \textcolor{black}{For a fair experimental evaluation, we conduct independent training and evaluation of our method and the comparison alternatives on the KAIST \cite{hwang2015multispectral}, RoadScene \cite{xu2020u2fusion}, MFNet \cite{ha2017mfnet}, and M3FD \cite{liu2022target} datasets, using the training and testing sets introduced above.} The epoch of training process is set to 200. The batch size is 16 and the initial learning rate is set to 0.0001 with exponential decay. The code of our method is implemented in Pytorch 1.7.1 framework with CUDA 10.1 and cuDNN 7.6.5 libraries, \textcolor{black}{and all experiments are implemented on a PC with a 2.40 GHz Intel Xeon E5-2680 v4 PCU and a NVIDIA Geforce GTX 2080ti GPU.}


\textbf{Evaluation Metrics:} We select the two commonly-used metrics, i.e., the average endpoint error (AEPE) and the percentage of correct keypoint (PCK), to evaluate the accuracy of optical flow estimation. AEPE is defined as the average Euclidean distance between the estimated and GT optical flow maps. PCK reflects the percentage of the correctly matched estimated points that are within a certain pixel-distance threshold from the corresponding GT points. In addition, six metrics, including CC, NCC, MI, peak signal-to-noise ratio (PSNR), the sum of the correlations of differences (SCD), and structural similarity index measure (SSIM), are used to measure the correlation and similarity between the predicted and the GT warped visible images.

\begin{table*}[tb]
	\small
	\renewcommand{\arraystretch}{1.2}
	\centering
	\caption{\textcolor{black}{Quantitative comparison to the SOTA methods on the MFNet dataset \cite{ha2017mfnet}. The best and second-best results are marked in \textbf{bold} and \underline{underlined} respectively. ``$\mathbf{\uparrow}$'' denotes the higher the better, ``$\mathbf{\downarrow}$'' denotes the lower the better.}}
	\resizebox{0.91\textwidth}{!}{
		\begin{tabular}{cccccccccc}
			\hline
			\multicolumn{2}{c}{Metrics} &
			\multicolumn{1}{c}{NTG \cite{chen2017normalized}} & 
			\multicolumn{1}{c}{SCB \cite{cao2020boosting}} &
			\multicolumn{1}{c}{DGC-Net \cite{melekhov2019dgc}} &
			\multicolumn{1}{c}{NeMAR \cite{arar2020unsupervised}} &
			\multicolumn{1}{c}{GMFlow \cite{xu2022gmflow}} &
			\multicolumn{1}{c}{CMF \cite{zhou2022promoting}} &
			\multicolumn{1}{c}{UMF-CMGR \cite{wang2022unsupervised}}&
			\multicolumn{1}{c}{SC$^{3}$EF (ours)} \\
			\hline
			\multicolumn{2}{c}{AEPE $\mathbf{\downarrow}$} & 312.58 & 8004.43 
			& 26.97 
			& 32.10 & 23.39 & \underline{17.80} & 34.33 & \textbf{6.76} \\
			\multirow{3}{*}{PCK ($\%$) $\mathbf{\uparrow}$} & 
			\multicolumn{1}{c}{$1 px$} & 1.57 & 0.94 & 3.61 & 3.11 & 4.89 
			& \underline{11.52} & 3.11 & \textbf{16.02} \\
			~ & $3 px$ & 4.67 & 2.81 & 10.81 & 9.31 & 14.55 & \underline{32.39} & 9.30 & \textbf{45.98} \\
			~ & $5 px$ & 7.71 & 4.67 & 17.95 & 15.38 & 23.78 & \underline{46.77} & 15.36 & \textbf{68.64} \\
			\multicolumn{2}{l}{CC $\mathbf{\uparrow}$} & 0.51 & 0.53 & 
			0.72 & 0.70 & \underline{0.75} & 0.79 & 0.70 & \textbf{0.86} \\
			\multicolumn{2}{c}{NCC $\mathbf{\uparrow}$} & 0.08 & 0.09 & 0.12 & 0.12 & 0.15 & 
			\underline{0.17} & 0.12 & \textbf{0.21} \\
			\multicolumn{2}{c}{MI $\mathbf{\uparrow}$} & 5.21 & 4.39 & \underline{7.09} & 7.23 & 
			6.73 & 6.91 & \textbf{7.24} & 6.68 \\
			\multicolumn{2}{c}{PSNR $\mathbf{\uparrow}$} & 9.87 & 9.94 & 13.10 & 12.77 & 13.22 & 
			\underline{13.80} & 12.77 & \textbf{16.16} \\
			\multicolumn{2}{c}{SCD $\mathbf{\uparrow}$} & 0.11 & 0.20 & 0.37 & 0.35 & 0.46 & 
			\underline{0.53} & 0.35 & \textbf{0.70} \\
			\multicolumn{2}{c}{SSIM $\mathbf{\uparrow}$} & 0.32 & 0.29 & 0.45 & 0.45 & 0.49 & 
			\underline{0.52} & 0.44 & \textbf{0.61} \\
			\hline
	\end{tabular}}
	\label{mfnet_com}
\end{table*}

\begin{figure*}[tb]
	\normalsize
	\centering
	\includegraphics[scale=0.41]{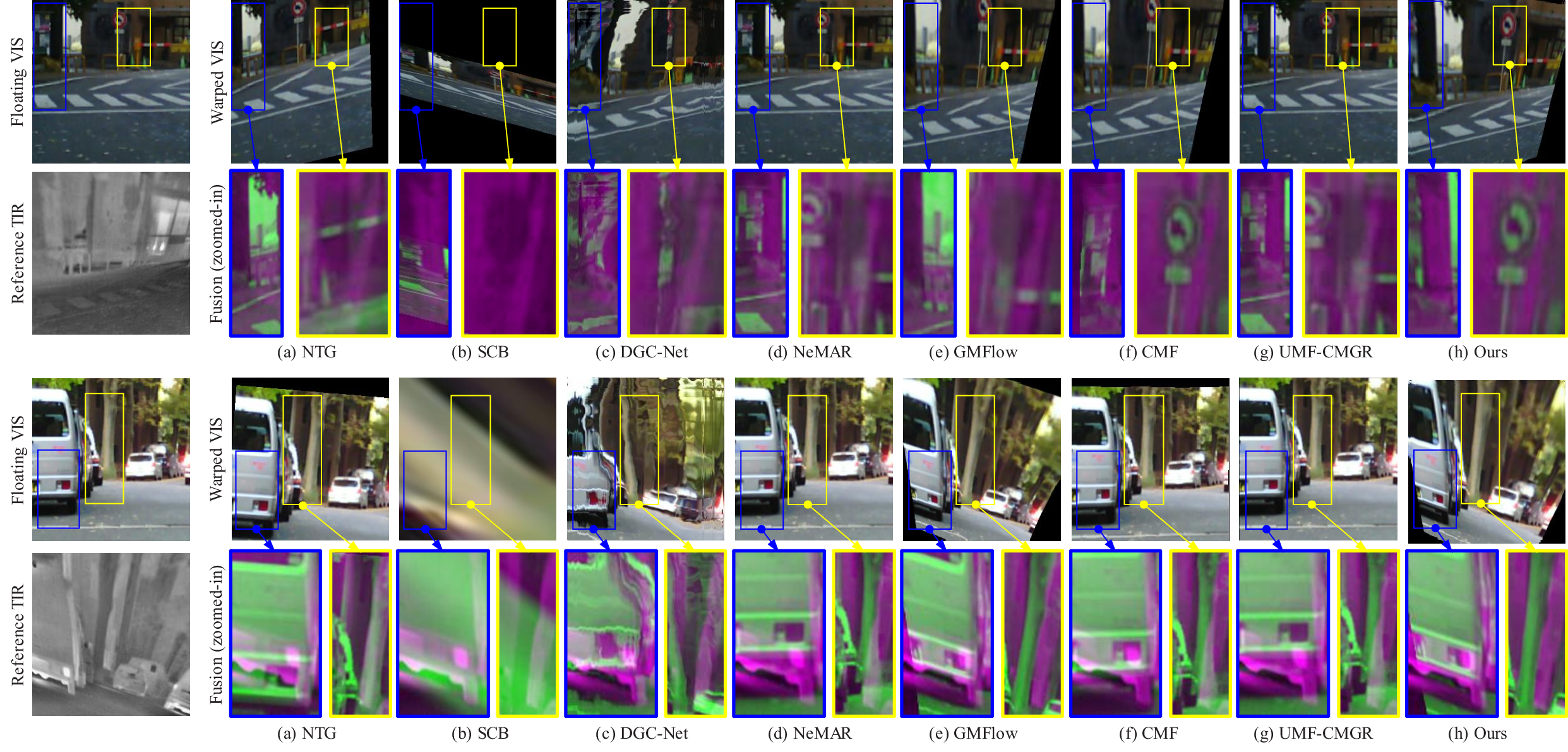}
	\caption{\textcolor{black}{Visual comparison of SOTA methods and SC$^{3}$EF on the MFNet dataset \cite{ha2017mfnet}. The warped visible images, paired with their corresponding zoomed-in fusion results highlighted in yellow and blue boxes, are presented to visualize the misalignments.}}
	\label{mfnet_res}
\end{figure*}

\subsection{Comparison with SOTA methods}
We perform both quantitative and qualitative experiments on the KAIST and RoadScene datasets to compare our proposed SC$^{3}$EF with two traditional cross-modality image registration methods, including NTG \cite{chen2017normalized} and SCB \cite{cao2020boosting}, and five SOTA deep learning based methods, including DGC-Net \cite{melekhov2019dgc}, NeMAR \cite{arar2020unsupervised}, GMFlow \cite{xu2022gmflow}, CMF \cite{zhou2022promoting}, and UMF-CMGR \cite{wang2022unsupervised}.

\begin{table*}[tb]
	\small
	\renewcommand{\arraystretch}{1.2}
	\centering
	\caption{\textcolor{black}{Quantitative comparison to the 
			SOTA methods on the M3FD dataset 
			\cite{liu2022target}. The best and second-best results are marked in \textbf{bold} and \underline{underlined} respectively. ``$\mathbf{\uparrow}$'' denotes the higher the better, ``$\mathbf{\downarrow}$'' denotes the lower the better.}}
	\resizebox{0.91\textwidth}{!}{
		\begin{tabular}{cccccccccc}
			\hline
			\multicolumn{2}{c}{Metrics} &
			\multicolumn{1}{c}{NTG \cite{chen2017normalized}} & 
			\multicolumn{1}{c}{SCB \cite{cao2020boosting}} &
			\multicolumn{1}{c}{DGC-Net \cite{melekhov2019dgc}} &
			\multicolumn{1}{c}{NeMAR \cite{arar2020unsupervised}} &
			\multicolumn{1}{c}{GMFlow \cite{xu2022gmflow}} &
			\multicolumn{1}{c}{CMF \cite{zhou2022promoting}} &
			\multicolumn{1}{c}{UMF-CMGR \cite{wang2022unsupervised}}&
			\multicolumn{1}{c}{SC$^{3}$EF (ours)} \\
			\hline
			\multicolumn{2}{c}{AEPE $\mathbf{\downarrow}$} & 243.59 & 109.78 & 
			20.01 & 32.10 & \underline{12.77} & 17.15 & 34.33 & \textbf{9.21} \\
			\multirow{3}{*}{PCK ($\%$) $\mathbf{\uparrow}$} & 
			\multicolumn{1}{c}{$1 px$} & 1.52 & 1.90 & 6.91 & 3.11 & 10.79 & 
			\textbf{13.86} & 3.11 & \underline{13.62} \\
			~ & $3 px$ & 4.64 & 5.56 & 20.12 & 9.31 & 30.04 & \underline{36.27} & 9.30 & 
			\textbf{38.78} \\
			~ & $5 px$ & 7.70 & 9.38 & 31.93 & 15.38 & 45.34 & \underline{50.70} & 15.36 & 
			\textbf{58.93} \\
			\multicolumn{2}{c}{CC $\mathbf{\uparrow}$} & 0.57 & 0.75 & 
			0.81 & 0.77 & 0.86 & \underline{0.83} & 0.77 & \textbf{0.88} \\
			\multicolumn{2}{c}{NCC $\mathbf{\uparrow}$} & 0.07 & 0.14 & 
			0.15 & 0.13 & \underline{0.18} & 0.18 & 0.13 & \textbf{0.20} \\
			\multicolumn{2}{c}{MI $\mathbf{\uparrow}$} & 3.90 & 5.18 & 
			\underline{5.95} & 6.04 & 5.61 & 5.78 & \textbf{6.07} & 5.68 \\
			\multicolumn{2}{c}{PSNR $\mathbf{\uparrow}$} & 9.57 & 11.96 & 
			\underline{15.94} & 15.40 & 15.17 & 13.91 & 15.32 & \textbf{16.19} \\
			\multicolumn{2}{c}{SCD $\mathbf{\uparrow}$} & 0.11 & 0.29 & 
			0.41 & 0.34 & \underline{0.54} & 0.44 & 0.34 & \textbf{0.59} \\
			\multicolumn{2}{c}{SSIM $\mathbf{\uparrow}$} & 0.39 & 0.50 & 
			0.58 & 0.57 & \underline{0.62} & 0.560 & 0.56 & \textbf{0.65} \\
			\hline
	\end{tabular}}
	\label{m3fd_com}
\end{table*}

\begin{figure*}[tb]
	\normalsize
	\centering
	\includegraphics[scale=0.41]{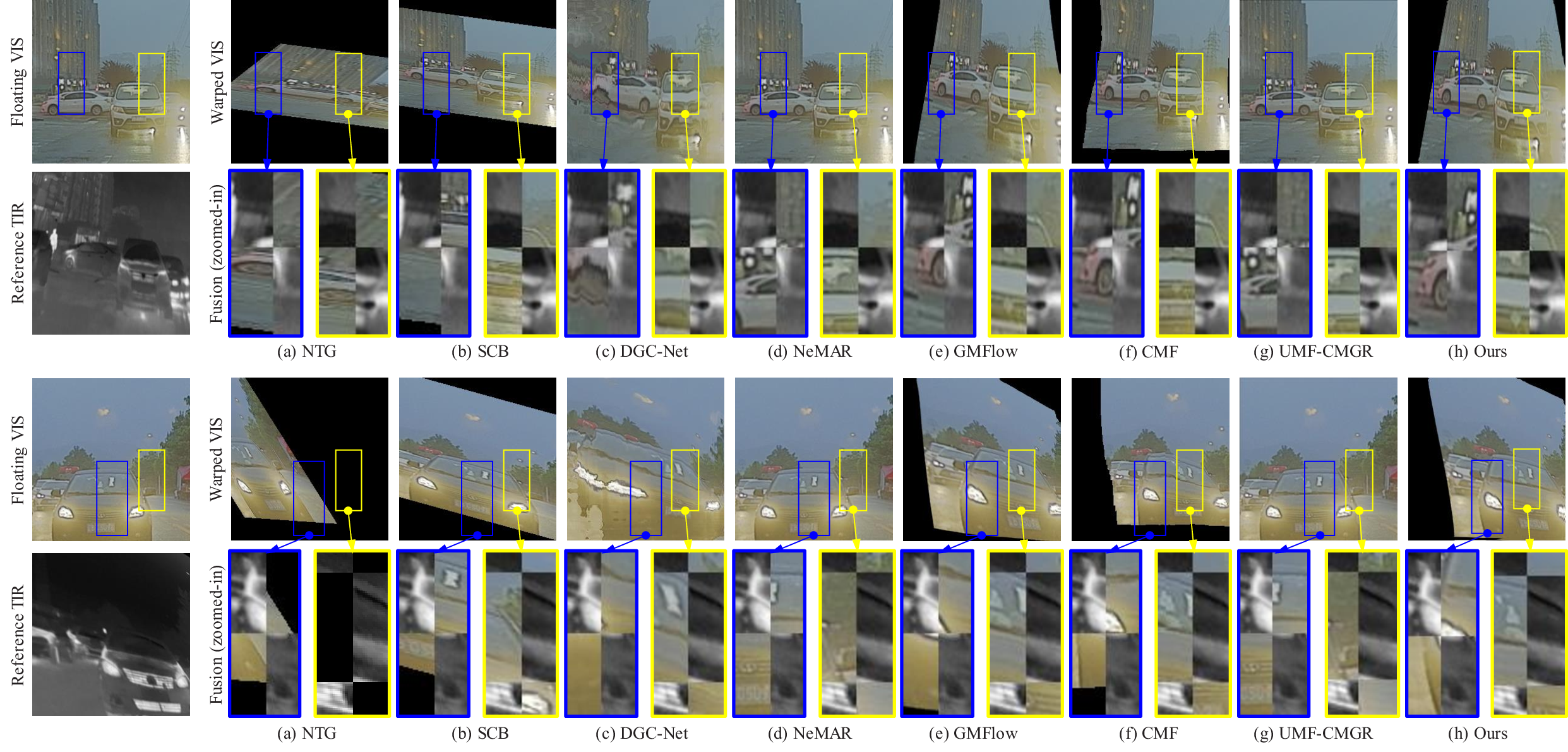}
	\caption{\textcolor{black}{Visual comparison of SOTA methods and SC$^{3}$EF on the M3FD dataset \cite{liu2022target}. The warped visible images, paired with their corresponding zoomed-in fusion results highlighted in yellow and blue boxes, are presented to visualize the misalignments.}}
	\label{m3fd_res}
\end{figure*}

\subsubsection{Experiments on KAIST dataset}
\label{exp_kaist}

We first evaluate the performance of different image registration methods on the KAIST dataset. The quantitative evaluation results (AEPE, PCK, CC, NCC, MI, PSNR, SCD, and SSIM) of our SC$^{3}$EF and the other image registration approaches are shown in Table~\ref{kaist_com}. It is noted that SC$^{3}$EF can outperform other alternatives by large margins, almost achieving the best scores in all evaluation metrics. In addition, it is worth mentioning that the AEPE score of our method is nearly three times better than the second best performer, which can directly reflect the accuracy of correspondence estimation. GMFlow and CMF perform almost equally well as the second best. The performance of DGC-Net lies in the middle level among the seven compared methods. The NTG, SCB, NeMAR and UMF-CMGR algorithms based on affine transformation models perform dramatically worse than others.

Some visual comparison examples are shown in Fig.~\ref{kaist_res}. It is observed that NTG, SCB, NeMAR and UMF-CMGR cannot achieve satisfactory results for salient foreground object registration, as highlighted in yellow and blue boxes. Even though DGC-Net can decently align visible and thermal images, it incurs some undesirable artifacts. While GMFlow and CMF perform better than the above-mentioned methods and can generally capture short-range and long-range correspondences simultaneously, they still cannot achieve accurate registration for visible and thermal images with large displacements. In comparison, our proposed SC$^{3}$EF is capable of estimating large displacements, achieving more accurate registration, and avoiding unsatisfactory artifacts. The registered visible and thermal images can be further utilized to facilitate other low-level and high-level visual computing applications in multispectral systems.

\subsubsection{Experiments on RoadScene dataset}
\label{exp_road}

We further evaluate the effectiveness of our proposed SC$^{3}$EF on the relatively smaller but more challenging RoadScene dataset, which contains a large proportion of homogeneous image regions. As shown in Table~\ref{road_com}, our method still achieves significantly higher accuracy than other alternatives for dense correspondence estimation. Some visual comparison examples are shown in Fig.~\ref{kaist_res}. It is observed that the traditional NTG and SCB methods and CNN-based DGC-Net, NeMAR, and UMF-CMGR models cannot generate correct correspondences, as highlighted in the yellow and blue boxes. While GMFlow and CMF generally perform better than the above-mentioned methods, they cannot achieve accurate registration results for indistinctive objects such as the road lane highlighted in the yellow box. In comparison, the proposed SC$^{3}$EF model can still achieve satisfactory registration results, even in these more challenging cases.

\subsubsection{\textcolor{black}{Experiments on MFNet dataset}}
\label{exp_mfnet}

\textcolor{black}{We further conduct the quantitative and qualitative evaluations on the MFNet dataset, which includes some low-light conditions. The comparative results of our proposed SC3EF and the other methods are shown in Table~\ref{mfnet_com}. It is noted that SC$^{3}$EF can still outperform other alternatives by large margins, almost achieving the best scores in all evaluation metrics.}

\textcolor{black}{Some visual comparison examples are shown in Fig.~\ref{mfnet_res}. It can be observed that other alternatives still struggle to achieve accurate registration for RGB-T images characterized by significant modal discrepancies and spatial displacements. In comparison, by combining HF local features with LF global self-correlations, our proposed SC$^{3}$EF model can effectively capture and correlate with temperature-distinguished objects across RGB-T modalities, such as guideposts, vehicles, and barriers.}

\begin{table*}[tb]
	\small
	\renewcommand{\arraystretch}{1.25}
	\centering
	\caption{\textcolor{black}{Paired t-test (t-statistic and P($|T| \leq |t|$)) comparison of each evaluation metric between representative cross-modal image registration methods and the proposed SC$^{3}$EF on the KAIST dataset \cite{hwang2015multispectral}. ``\checkmark'' denotes that the proposed SC$^{3}$EF achieves a statistically significant improvement (P$<$0.05) across all metrics compared to the alternative method.}}
	
	\resizebox{0.79\textwidth}{!}{
		\begin{tabular}{cccccccccc}
			\hline
			\multicolumn{2}{c}{\multirow{2}{*}{Metrics}} &
			\multicolumn{2}{c}{NeMAR-Ours} & 
			\multicolumn{2}{c}{GMFlow-Ours} & 
			\multicolumn{2}{c}{CMF-Ours} & 
			\multicolumn{2}{c}{UMF-CMGR-Ours} \\
			\cline{3-10}
			\multicolumn{2}{c}{} & \multicolumn{1}{c}{$t$-Stat} & P($|T|\leq|t|$) & \multicolumn{1}{c}{$t$-Stat} & P($|T|\leq|t|$) & \multicolumn{1}{c}{$t$-Stat} & P($|T|\leq|t|$) & \multicolumn{1}{c}{$t$-Stat} & P($|T|\leq|t|$) \\
			\hline
			\multicolumn{2}{c}{AEPE} & 71.66 & 0.00 & 34.76 & 3.79e-174 & 28.36 & 2.76e-130 & 62.83 & 0.00 \\
			\multirow{3}{*}{PCK} & 1px & -58.52 & 0.00 & -35.06 & 3.49e-176 & -22.86 & 2.14e-93 & -56.57 & 9.37e-314 \\
			& 3px & -86.70 & 0.00 & -48.13 & 1.80e-262 & -33.44 & 3.99e-165 & -82.33 & 0.00 \\
			& 5px & -122.57 & 0.00 & -54.70 & 1.01e-302 & -41.00 & 2.64e-216 & -110.53 & 0.00 \\
			\multicolumn{2}{c}{CC} & -49.77 & 9.64e-273 & -31.48 & 4.46e-123 & -23.27 & 7.95e-97 & -40.10 & 2.27e-200 \\
			\multicolumn{2}{c}{NCC} & -80.78 & 0.00 & -49.24 & 1.88e-269 & -35.34 & 2.44e-142 & -72.73 & 0.00 \\
			\multicolumn{2}{c}{MI} & 20.70 & 1.85e-79 & -12.21 & 4.46e-32 & -4.49 & 9.75e-6 & 31.61 & 1.54e-157 \\
			\multicolumn{2}{c}{PSNR} & -28.93 & 7.88e-144 & -18.05 & 1.41e-68 & -14.22 & 1.09e-45 & -29.62 & 5.84e-151 \\
			\multicolumn{2}{c}{SCD} & -46.79 & 5.70e-254 & -36.48 & 6.36e-188 & -23.66 & 7.36e-99 & -46.84 & 6.94e-255 \\
			\multicolumn{2}{c}{SSIM} & -37.27 & 3.01e-191 & -27.99 & 1.20e-121 & -16.84 & 3.24e-58 & -36.91 & 6.69e-187 \\
			\hline
			\multicolumn{2}{c}{Average} & -41.90 & 1.85e-89 & -31.48 & 4.46e-33 & -23.27 & 7.95e-7 & -56.91 & 6.39e-111 \\
			\multicolumn{2}{c}{\textbf{P$<$0.05}} & \multicolumn{2}{c}{\checkmark} &
			\multicolumn{2}{c}{\checkmark} &
			\multicolumn{2}{c}{\checkmark} &
			\multicolumn{2}{c}{\checkmark} \\
			\hline
		\end{tabular}
	}
	\label{paired_ttest}
\end{table*}

\begin{figure*}[tb]
	\normalsize
	\centering
	\includegraphics[scale=0.35]{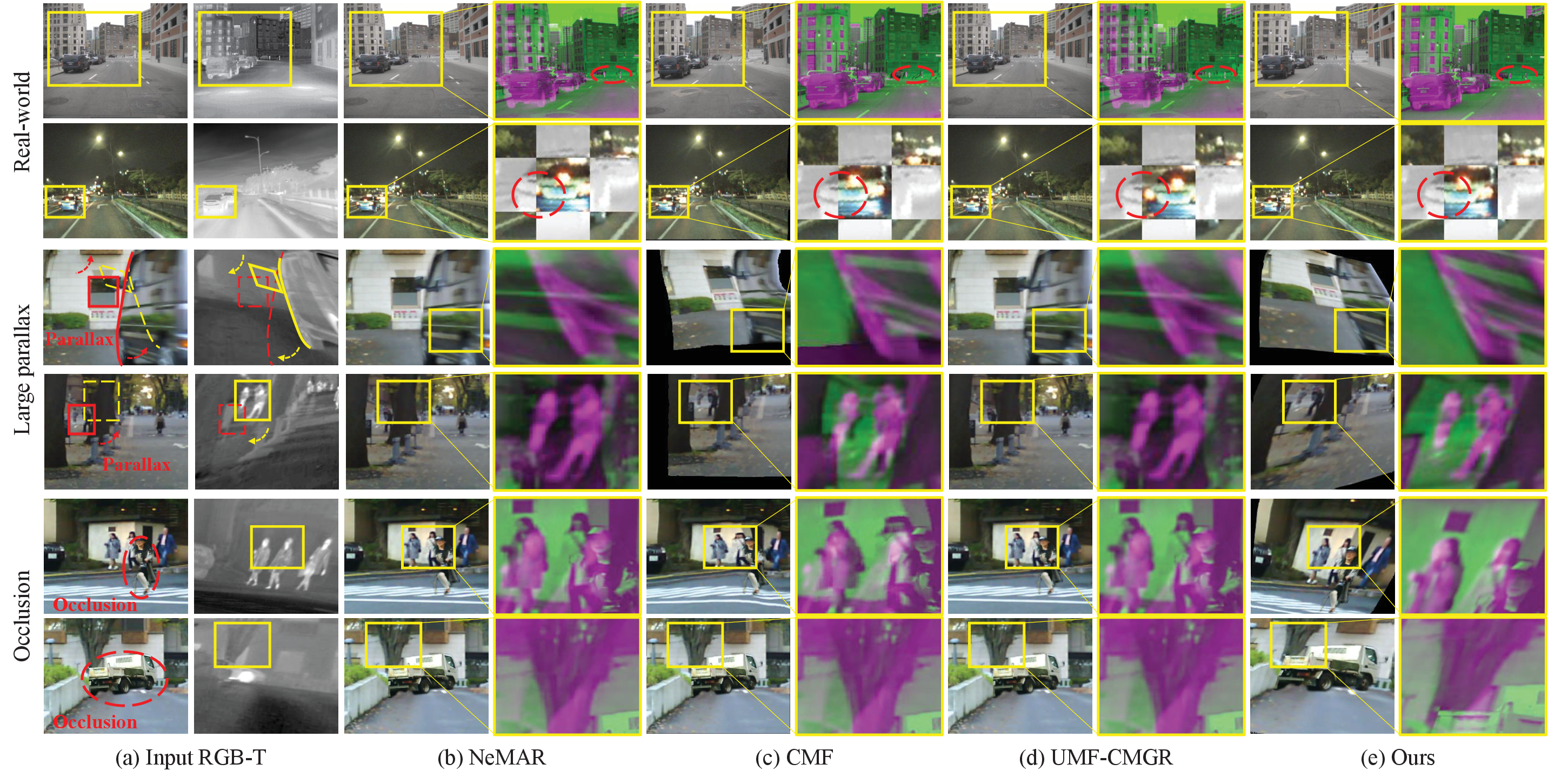}
	\caption{\textcolor{black}{Visual comparison of representative cross-modal image registration methods and SC$^{3}$EF on real-world traffic scenes (first two rows), simulated large parallax (middle two rows) and occluded conditions (last two rows). The warped visible images, paired with their corresponding zoomed-in fusion results, are presented to visualize the misalignments.}}
	\label{real_plus_occlusion}
\end{figure*}

\begin{table*}[tb]
	\small
	\renewcommand{\arraystretch}{1.25}
	\centering
	\caption{\textcolor{black}{
			Quantitative comparison of AEPE and PCK-3px ($\%$) metrics between SOTA methods and ours on different adverse weather subsets on the M3FD \cite{liu2022target} dataset. The best and second-best results are marked in \textbf{bold} and \underline{underlined} respectively.}}
	\resizebox{0.84\textwidth}{!}{
		\begin{tabular}{ccccccccccc}
			\hline
			\multicolumn{1}{c}{Conditions} & 
			\multicolumn{2}{c}{Normal-light} &
			\multicolumn{2}{c}{Low-light} & 
			\multicolumn{2}{c}{Extreme low-light} &
			\multicolumn{2}{c}{Rainy} &
			\multicolumn{2}{c}{Hazy} \\
			\hline
			Metrics & AEPE & PCK-3\textit{px} & AEPE & PCK-3\textit{px}& AEPE & PCK-3\textit{px}& AEPE & PCK-3\textit{px} & AEPE & PCK-3\textit{px} \\
			\hline
			NTG & 198.06 & 4.46 & 241.91 & 5.32 & 169.42 & 2.53 & 138.01 & 5.21 & 206.28 & 4.47 \\
			SCB & 70.30 & 6.04 & 105.46 & 4.87 & 111.88 & 7.28 & 83.52 & 6.58 & 125.57 & 3.62 \\
			DGC-Net & 20.25 & 19.88 & 22.16 & 17.63 & 12.80 & \underline{25.51} & 19.70 & 18.83 & 9.77 & 34.68  \\
			NeMAR & 32.09 & 9.27 & 32.13 & 9.30 & 31.63 & 10.43 & 30.89 & 9.89 & 34.51 & 8.47 \\
			GMFlow & \underline{10.94} & 33.61 & \underline{15.62} & 24.89 & \underline{12.01} & 25.042 & \underline{11.44} & 25.643 & \underline{9.03} & \underline{38.17} \\
			CMF & 13.77 & \textbf{42.89} & 19.03 & \underline{31.57} & 24.82 & 22.69 & 15.94 & \underline{35.19} & 20.464 & 30.01 \\
			UMF-CMGR & 34.33 & 9.41 & 34.35 & 9.07 & 33.82 & 9.78 & 33.02 & 8.82 & 36.91 & 8.44 \\
			SC$^{3}$EF & \textbf{8.44} & \underline{40.36} & \textbf{11.07} & \textbf{34.26} & \textbf{7.84} & \textbf{40.20} & \textbf{6.37} & \textbf{45.99} & \textbf{6.62} & \textbf{51.55} \\
			\hline
	\end{tabular}}
	\label{abla_weather}
\end{table*}

\begin{figure*}[tb]
	\normalsize
	\centering
	\includegraphics[scale=0.34]{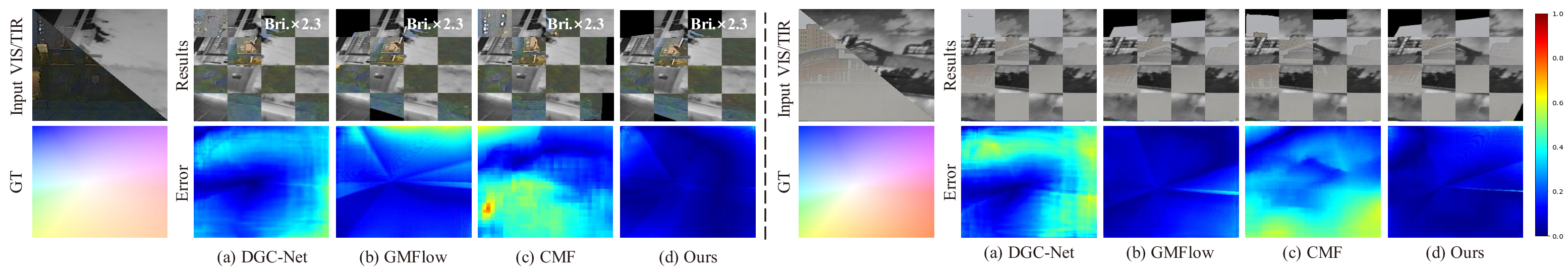}
	\caption{\textcolor{black}{Visual comparison with representative cross-modal image registration methods on the M3FD dataset \cite{liu2022target} under adverse environmental conditions. The first row shows the input images and the corresponding warped results, while the second row displays the GT optical flow maps and their corresponding error visualizations.}}
	\label{m3fd_weather_ablation}
\end{figure*}

\subsubsection{\textcolor{black}{Experiments on M3FD dataset}}
\label{exp_m3fd}

\textcolor{black}{To validate the advantages of thermal images under different weather conditions, we also conduct experiments on the M3FD dataset, which includes diverse weather conditions such as extreme low-light, rain, and haze. As shown in Table~\ref{m3fd_com}, our method still achieves significantly higher accuracy than other alternatives for dense correspondence estimation.}

\textcolor{black}{Some visual comparison examples are shown in Fig.~\ref{m3fd_res}. It is observed that the comparative methods cannot achieve satisfactory registration results for large displacement and low-light conditions, as highlighted in yellow and blue boxes. While GMFlow can decently align visible and thermal images, it still incurs some undesired artifacts. In contrast, our proposed HF/LF component decoupling encoding strategy can reduce the difficulty of feature extraction from degraded images effectively. The proposed SC$^{3}$EF extracts local salient features from the HF components and constructs global self-correlations from the LF components, demonstrating competitive registration performance even on low-quality RGB-T images captured under adverse weather conditions.}

\subsubsection{\textcolor{black}{Experiments of statistical significance}}
\label{significance}

\textcolor{black}{We also conduct paired t-tesets \cite{hsu2014paired} on the KAIST dataset \cite{hwang2015multispectral} across the abovementioned metrics to evaluate the statistical significance of our proposed SC$^{3}$EF by comparing with four representative cross-modal image registration methods (NeMAR \cite{arar2020unsupervised}, GMFlow \cite{xu2022gmflow}, CMF \cite{zhou2022promoting}, and UMF-CMGR \cite{wang2022unsupervised}). As depicted in Table \ref{paired_ttest}, our proposed SC$^{3}$EF achieves statistically significant improvements (P$<$0.05) over other compared methods across all metrics. It confirms the robustness and effectiveness of SC$^{3}$EF, proving its capability to provide accurate registration and consistent performance compared to the SOTA methods.}

\begin{figure*}[tb]
	\normalsize
	\centering
	\includegraphics[scale=0.28]{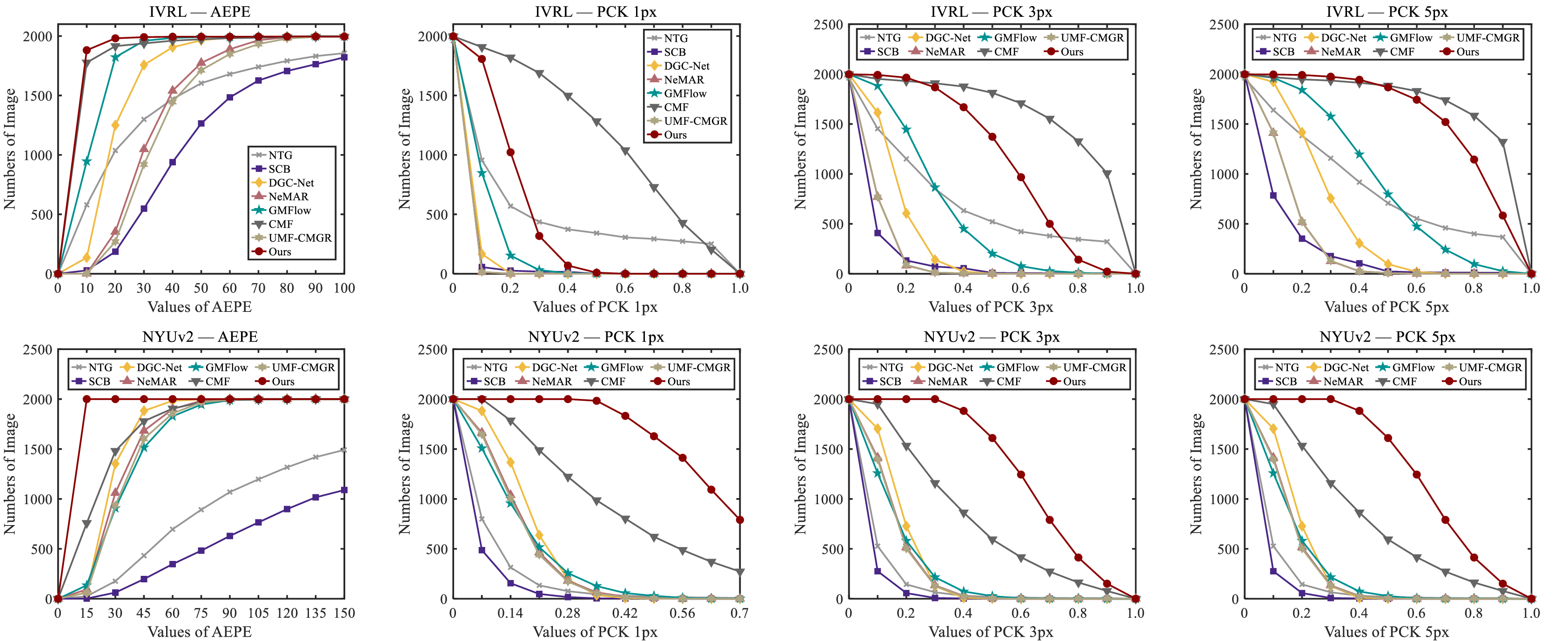}
	\caption{\textcolor{black}{Quantitative comparison of AEPE and PCK metrics with SOTA methods on the IVRL \cite{brown2011multi} and NYUv2 \cite{silberman2012indoor} datasets. In the first column, each point represents the number of images with AEPE values less than the corresponding x-axis value, while in the other plots, it represents the number of images with PCK values greater than the corresponding x-axis value.}}
	\label{ivrl_nyuv2_metrics}
\end{figure*}

\begin{figure*}[tb]
	\normalsize
	\centering
	\includegraphics[scale=0.34]{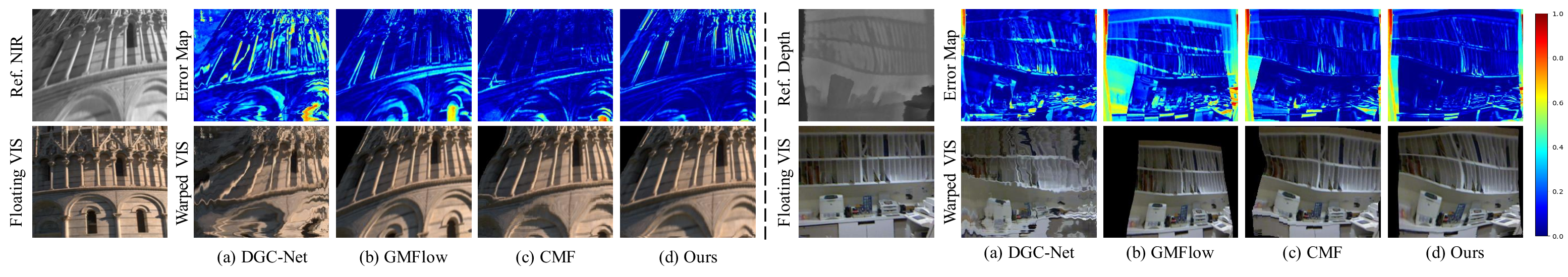}
	\caption{\textcolor{black}{Visual comparison with representative cross-modal image registration methods on the IVRL \cite{brown2011multi}  (first five columns) and NYUv2 \cite{silberman2012indoor} (last five columns) datasets. The first row presents the error maps between the warped results and the GTs. The second row displays the warped visible images generated by each method.}}
	\label{ivrl_nyuv2_res}
\end{figure*}

\subsection{\textcolor{black}{Generalization experiments}}
\label{generalization}

\subsubsection{\textcolor{black}{Experiments on real-world misalignment data}}
\label{real_dataset}

\textcolor{black}{We further perform the generalization experiments on 100 pairs of real traffic scene images, with 50 pairs from the  FLIR ADAS dataset\footnote{https://www.flir.com/oem/adas/adas-dataset-form/} and 50 pairs from our own collected data. Some visual comparisons of the registration performances in the daytime and nightime real-world scenes are shown in the first two rows of Fig. \ref{real_plus_occlusion}. The warped visible images alongside  their corresponding zoomed-in fusion results are presented to visualize the misalignments. It is noted that CMF \cite{zhou2022promoting} achieves more accurate cross-modal correspondence estimation compared to NeMAR \cite{arar2020unsupervised} and UMF-CMGR \cite{wang2022unsupervised}. However, CMF \cite{zhou2022promoting} still struggles to accurately align the distant fine-grained objects, such as pedestrians and buildings. In comparison, our proposed SC$^{3}$EF demonstrates a greater ability to accurately estimate correspondences for detailed features. This highlights its generalization capability and practicality for RGB-T image registration in real-world traffic scenarios.}

\subsubsection{\textcolor{black}{Performance on large parallax and severe occlusion scenes}}
\label{ablation_occlution}

\textcolor{black}{We conduct additional generalization experiments on the MFNet dataset \cite{ha2017mfnet}, focusing on simulated  scenarios with large parallax and severe occlusion. Specifically, we generate large parallax RGB-T image pairs using the data synthesis strategy introduced in Section \ref{exp_setting}, based on different transformation models (Aff, HG, and TPS). To simulate occlusions, we utilize the segmentation annotations from the MFNet \cite{ha2017mfnet} dataset to extract individual human and vehicle objects, which are then randomly overlaid onto the visible images. Based on the simulated images with large parallax and severe occlusion, qualitative comparison of three representative cross-modal image registration methods (NeMAR \cite{arar2020unsupervised}, CMF \cite{zhou2022promoting}, and UMF-CMGR \cite{wang2022unsupervised}) and the proposed SC$^{3}$EF are illustrated in the last four rows of Fig. \ref{real_plus_occlusion}. The parallax between the input RGB-T images is visualized in column (a) of the middle two rows using red and yellow contour lines in visible and thermal images, respectively. The simulated occluded objects are highlighted using red dashed boxes in column (a) of the last two rows. The warped visible images paired with their corresponding zoomed-in fusion results are presented to visualize image misalignments. It is observed that our proposed SC$^{3}$EF achieves more accurate correspondence estimation and demonstrates stronger adaptability and robustness in both large parallax and severe occluded scenarios. In comparison, other alternatives are more affected by the large displacements and occlusions in the source images, leading to considerable interference in their cross-modal correspondence estimations.}

\subsubsection{\textcolor{black}{Performance on different adverse weather conditions}}
\label{ablation_weather}
\begin{table}[tb]
	\Large
	\renewcommand{\arraystretch}{1.25}
	\centering
	\caption{\textcolor{black}{Computational efficiency comparison with deep learning-based SOTA methods on the KAIST \cite{hwang2015multispectral}, RoadScene \cite{xu2020u2fusion} and MFNet \cite{ha2017mfnet} datasets. \underline{\textbf{Bold and underlined}} indicates the best, \textbf{bold} indicates the second-best, and \underline{underlined} indicates the third-best, respectively. ``$\mathbf{\uparrow}$'' denotes the higher the better, ``$\mathbf{\downarrow}$'' denotes the lower the better.}}
	\resizebox{0.48 \textwidth}{!}{
		\begin{tabular}{ccccccc}
			\hline
			\multicolumn{1}{c}{Methods} & 
			\multicolumn{2}{c}{KAIST \cite{hwang2015multispectral}} &
			\multicolumn{2}{c}{RoadScene \cite{xu2020u2fusion}} &
			\multicolumn{2}{c}{MFNet \cite{ha2017mfnet}} \\
			\hline
			\multicolumn{1}{c}{Metrics} & \multicolumn{1}{c}{Latency (ms)$\mathbf{\downarrow}$} & \multicolumn{1}{c}{FPS$\mathbf{\uparrow}$} & \multicolumn{1}{c}{Latency (ms)$\mathbf{\downarrow}$} & \multicolumn{1}{c}{FPS$\mathbf{\uparrow}$} & \multicolumn{1}{c}{Latency (ms)$\mathbf{\downarrow}$} & \multicolumn{1}{c}{FPS$\mathbf{\uparrow}$} \\
			\hline
			\multicolumn{1}{c}{DGC-Net \cite{melekhov2019dgc}} & \multicolumn{1}{c}{134.42} & \multicolumn{1}{c}{7.44} & \multicolumn{1}{c}{89.85} & \multicolumn{1}{c}{11.13} & \multicolumn{1}{c}{90.75} & \multicolumn{1}{c}{11.02} \\
			\multicolumn{1}{c}{NeMAR \cite{arar2020unsupervised}} & \multicolumn{1}{c}{\underline{\textbf{57.40}}} & \multicolumn{1}{c}{\underline{\textbf{17.42}}} & \multicolumn{1}{c}{\underline{\textbf{31.84}}} & \multicolumn{1}{c}{\underline{\textbf{31.41}}} & \multicolumn{1}{c}{\underline{\textbf{55.91}}} & \multicolumn{1}{c}{\underline{\textbf{17.89}}} \\
			\multicolumn{1}{c}{GMFlow \cite{xu2022gmflow}} & \multicolumn{1}{c}{84.30} & \multicolumn{1}{c}{11.86} & \multicolumn{1}{c}{\textbf{47.53}} & \multicolumn{1}{c}{\underline{21.04}} & \multicolumn{1}{c}{84.06} & \multicolumn{1}{c}{11.90} \\
			\multicolumn{1}{c}{CMF \cite{zhou2022promoting}} & \multicolumn{1}{c}{104.09} & \multicolumn{1}{c}{9.61} & \multicolumn{1}{c}{63.77} & \multicolumn{1}{c}{15.68} & \multicolumn{1}{c}{100.02} & \multicolumn{1}{c}{10.00} \\
			\multicolumn{1}{c}{UMF-CMGR \cite{wang2022unsupervised}} & \multicolumn{1}{c}{\underline{81.76}} & \multicolumn{1}{c}{\underline{12.23}} & \multicolumn{1}{c}{\underline{46.45}} & \multicolumn{1}{c}{\textbf{21.53}} & \multicolumn{1}{c}{\textbf{75.79}} & \multicolumn{1}{c}{\textbf{13.19}} \\
			\multicolumn{1}{c}{Ours} & \multicolumn{1}{c}{\textbf{81.07}} & \multicolumn{1}{c}{\textbf{12.33}} & \multicolumn{1}{c}{80.00} & \multicolumn{1}{c}{12.50} & \multicolumn{1}{c}{\underline{79.65}} & \multicolumn{1}{c}{\underline{12.56}} \\
			\hline
	\end{tabular}}
	\label{efficiency_table}
\end{table}

\begin{table*}[tb]
	\large
	\renewcommand{\arraystretch}{1.25}
	\centering
	\caption{\textcolor{black}{Quantitative evaluation of the SC$^{3}$EF model on the KAIST \cite{hwang2015multispectral} dataset in terms of registration accuracy and computational efficiency, with a gradual deployment of the proposed GSCE, LFE, GCCE, and LCCE modules. The best results are marked in \textbf{bold}. The subscript values indicate the changes relative to the ``Case'' in the previous row. ``$\mathbf{\uparrow}$'' denotes the higher the better, ``$\mathbf{\downarrow}$'' denotes the lower the better.}}
	\resizebox{0.94\textwidth}{!}{
		\begin{tabular}{ccccc|cccccccc|cc}
			\hline
			\multirow{2}{*}{Case} & 
			\multirow{2}{*}{GSCE} &
			\multirow{2}{*}{LFE} & 
			\multirow{2}{*}{GCCE} &
			\multirow{2}{*}{LCCE} &
			\multirow{2}{*}{AEPE $\mathbf{\downarrow}$} &
			\multicolumn{3}{c}{PCK ($\%$) $\mathbf{\uparrow}$} &
			\multirow{2}{*}{CC $\mathbf{\uparrow}$} &
			\multirow{2}{*}{PSNR $\mathbf{\uparrow}$} &
			\multirow{2}{*}{SCD $\mathbf{\uparrow}$}&
			\multirow{2}{*}{SSIM $\mathbf{\uparrow}$} &
			\multirow{2}{*}{Latency (ms) $\mathbf{\downarrow}$} &
			\multirow{2}{*}{FPS $\mathbf{\uparrow}$} \\
			\cline{7-9} & ~ & ~ & ~ & ~ & ~ & $1 px$ & $3 px$ & $5 px$ & ~ & ~ & ~ & ~ & ~ & ~\\
			\hline
			\multicolumn{1}{c}{1} & $\checkmark$ & ~ & ~ & ~ & 14.38 & 7.78 & 23.00 & 37.24 & 0.80 & 14.17 & 0.54 & 0.57 & 31.06 & 32.20 \\
			\multicolumn{1}{c}{2} & ~ & $\checkmark$ & ~ & ~ & 12.16$_{+2.23}$ & 9.79 & 28.41 & 45.08 & 0.81 & 14.36 & 0.56 & 0.58 & \textbf{20.72$_{+\mathbf{10.34}}$} & \textbf{48.25$_{+\mathbf{16.05}}$} \\
			\multicolumn{1}{c}{3} & $\checkmark$ & $\checkmark$ & ~ & ~ & 10.27$_{+1.88}$ & 11.28 & 32.48 & 50.77 & 0.82 & 14.61 & 0.59 & 0.60 & 52.50$_{-31.78}$ & 19.65$_{-28.60}$ \\
			\multicolumn{1}{c}{4} & $\checkmark$ & $\checkmark$ & $\checkmark$ & ~ & 6.07$_{+\mathbf{4.21}}$ & 16.59 & 48.00 & 72.24 & 0.87 & 16.34 & 0.69 & 0.66 & 76.73$_{-24.23}$ & 13.03$_{-6.62}$ \\
			\multicolumn{1}{c}{5} & $\checkmark$ & $\checkmark$ & $\checkmark$ & $\checkmark$ & \textbf{4.65$_{+1.42}$} & \textbf{23.16} & \textbf{62.11} & \textbf{83.61} & \textbf{0.88} & \textbf{16.94} & \textbf{0.72} & \textbf{0.68} & 81.07$_{-4.34}$ & 12.33$_{-0.70}$ \\
			\hline
	\end{tabular}}
	\label{ablation_tab}
\end{table*}

\begin{figure*}[tb]
	\normalsize
	\centering
	\includegraphics[scale=0.43]{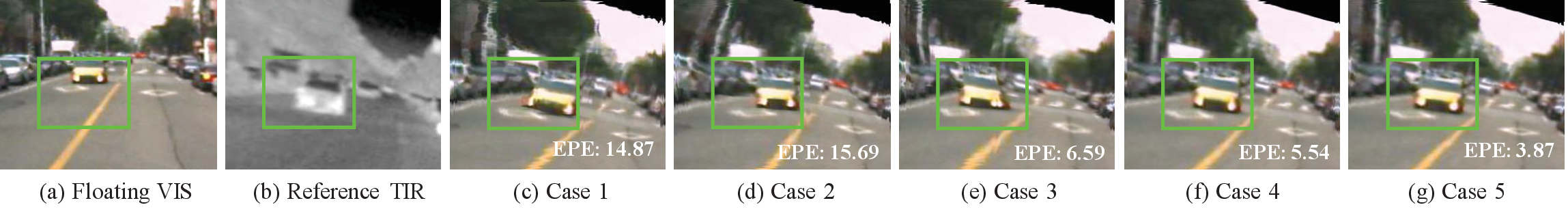}
	\caption{Ablation experiments to evaluate the effectiveness of the proposed GSCE, LFE, GCCE, and LCCE modules. (a) The floating visible images, (b) The reference thermal images, (c)-(g) The five ablation cases corresponding to the Table~\ref{ablation_tab}.}
	\label{ablation_fig}
\end{figure*}

\textcolor{black}{To validate the performance of our proposed SC$^{3}$EF under adverse environmental conditions, we make use of images captured under various extreme sensing scenarios in the M3FD dataset to conduct ablation experiments. Concretely, we categorize the test set of M3FD into four groups based on different lighting and weather conditions: low-light, extreme low-light, rainy, and hazy. As shown in the quantitative evaluation results in Table~\ref{abla_weather}, our proposed method achieves better or comparable performance across different environmental conditions. It is worth mentioning that our method demonstrated larger performance advantages in three extreme scenarios (e.g., extreme low-light, rainy, and hazy), proving that our SC$^{3}$EF can reliably extract and encode degraded visible features for cross-modal correspondence estimation.}

\textcolor{black}{As shown in Fig.~\ref{m3fd_weather_ablation}, some qualitative comparisons of the performances in extreme low-light and hazy scenes are depicted. The first row presents some representative examples from three recent deep learning-based methods including DGC-Net, GMFlow, CMF, and our SC$^{3}$EF. The second row shows visualizations of error maps calculated between the predicted optical flow and the ground truth optical flow. In both examples, when processing the homogeneous background or the salient foreground, our method achieves a more stable perception of RGB-T features and more accurate dense optical flow estimation across various extreme sensing conditions.}

\subsubsection{\textcolor{black}{Experiments on other cross-modal data}}
\label{cross_modal}

\textcolor{black}{We conduct generalization experiments on 120 pairs of IVRL (RGB-N) dataset \cite{brown2011multi} and 362 pairs of NYUv2 (RGB-D) dataset \cite{silberman2012indoor}, respectively.} To extensively perform evaluation, we randomly simulate and augment the testing dataset to 2000 pairs utilizing the same manner as depicted in Fig.~\ref{data_gen}.

\textcolor{black}{Different from RGB and thermal image pairs, RGB and NIR modalities present more similar characteristics. As shown in the quantitative comparison in the first row of Fig.~\ref{ivrl_nyuv2_metrics}, our proposed SC$^{3}$EF demonstrates competitive performance in terms of AEPE and PCK metrics, ranking second place after CMF. Some visual comparison examples are shown in the first five columns of Fig.~\ref{ivrl_nyuv2_res}. Compared to other alternatives, CMF and our method can provide more accurate correspondence estimations of the overall shape of the target, such as the outline of buildings. Notably, when applied to RGB-N modalities with similar image characteristics, the proposed SC$^{3}$EF model still delivers competitive generalization performance.}

\begin{figure}[tb]
	\normalsize
	\centering
	\includegraphics[scale=0.26]{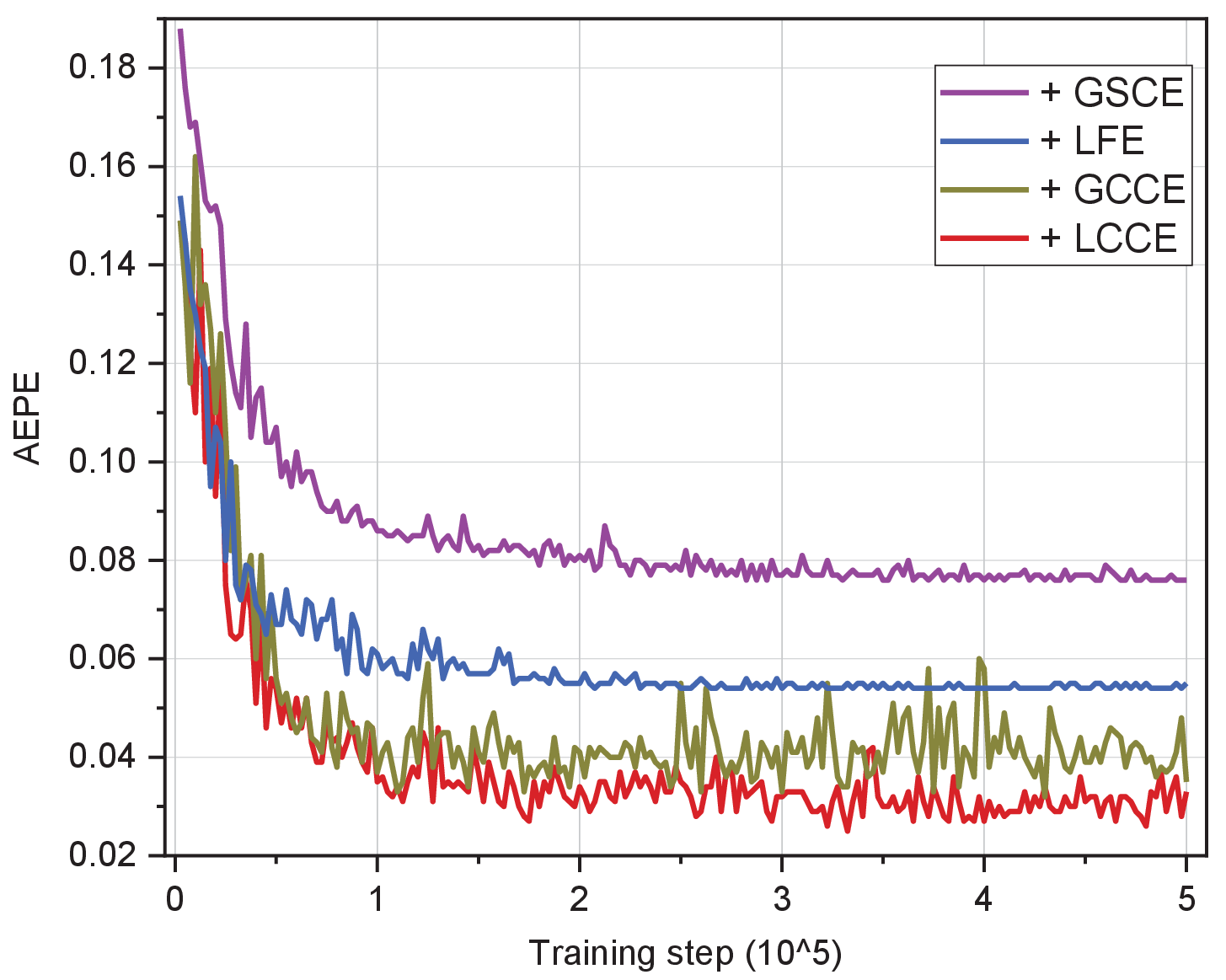}
	\caption{Ablation experiments to evaluate the effectiveness of the proposed GSCE, LFE, GCCE, and LCCE modules, achieving better testing performances in terms of AEPE metric in the training process.}
	\label{ablation_epe_fig}
\end{figure}

\textcolor{black}{The imaging principles of depth and visible images differ significantly, leading to considerable modality differences. As shown in the second row of Fig.~\ref{ivrl_nyuv2_metrics}, our method achieves significantly higher accuracy than other alternatives for dense correspondence estimation. Some visual comparison examples are shown in the last five columns of Fig.~\ref{ivrl_nyuv2_res}. It is observed that SOTA cross-modality image registration methods such as DGC-Net, GMFlow and CMF cannot generate satisfactory registration results. More specifically, they cannot capture precise correspondences for objects with extremely large modality discrepancies and displacements. In comparison, the proposed SC$^{3}$EF model shows obvious superiority and good generalization ability when applied to RGB-D images with significant modal discrepancies.}

\subsection{\textcolor{black}{Computational comparison}}
\label{efficiency}

\textcolor{black}{We conduct an additional evaluation to assess the computational efficiency of different deep learning-based SOTA image registration methods in terms of latency and frames per second (FPS) on KAIST \cite{hwang2015multispectral}, RoadScene \cite{xu2020u2fusion} and MFNet \cite{ha2017mfnet} datasets. Table \ref{efficiency_table} presents the average computational efficiency of all methods across different datasets. All the methods are tested using image pairs of full resolution to calculate the inference latency and FPS. Among the SOTA image registration methods, NeMAR \cite{arar2020unsupervised} achieves the lowest latency and highest FPS across all datasets. It is worth mentioning that our proposed SC$^{3}$EF delivers competitive performances and ranks among the top-3 on the KAIST \cite{hwang2015multispectral} and MFNet \cite{ha2017mfnet} datasets, verifying its computational efficiency through the proposed hybrid framework combining DWConv-based convolutions and PPM-based transformers.}

\subsection{Ablation study}
\label{ablation_study}

\subsubsection{\textcolor{black}{Effectiveness of proposed modules}}
\label{ablation_modules}

The SC$^{3}$EF is a convolution-transformer-based model consisting of a number of purpose-built modules for extracting distinctive local features, encoding global self-correlation cues, and estimating cross-modality correspondence based on the extracted local and global representations. In this section, we set up ablation experiments on the KAIST dataset to evaluate the \textcolor{black}{effectiveness and computational efficiency of each module we proposed}. In our experiments, we keep the hierarchical optical flow estimation stage unchanged and integrate the proposed GSCE, LFE, GCCE, and LCCE modules into the self-correlation extraction stage and the cross-correspondence estimation stage step by step. As presented in Table~\ref{ablation_tab} and Fig.~\ref{ablation_fig}, the experimental results validate the contributions of GSCE, LFE, GCCE, and LCCE modules, all leading to higher accuracy in dense correspondence estimation and better visual perception \textcolor{black}{by a decrease in computational efficiency}. Specifically, as shown in the first three cases in Table~\ref{ablation_tab} and Fig.~\ref{ablation_fig}, our proposed LFE and GSCE modules effectively extract local representative features and global contextual cues, both of which are important to establish correct and robust cross-correspondence estimation results for significantly different visible and thermal images. As shown in Case 4 and 5 in Table~\ref{ablation_tab} and Fig.~\ref{ablation_fig}, the proposed GCCE modules can make use of the encoded global self-correlations to handle large displacements, while LCCE modules can generate local feature-based matching results as auxiliary information to further refine the final correspondence estimation results. Furthermore, better testing performances are achieved in the training process by gradually adding GSCE, LFE, GCCE, and LCCE modules as shown in Fig.~\ref{ablation_epe_fig}, validating the effectiveness of these purpose-built modules. 

\begin{figure}[tb]
	\normalsize
	\centering
	\includegraphics[scale=0.28]{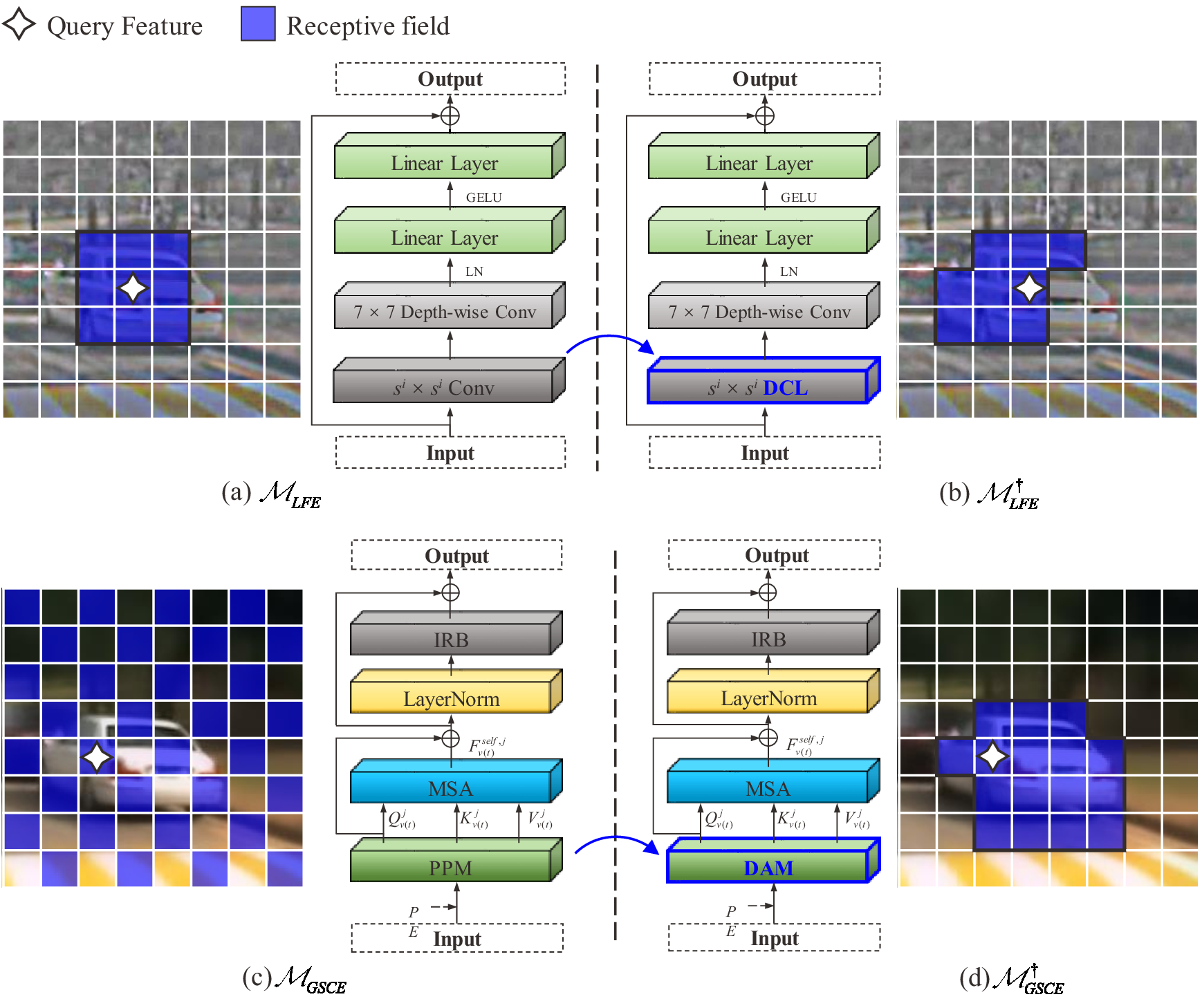}
	\caption{\textcolor{black}{Comparison of $\boldsymbol{\mathcal{M}}^{\dagger}_{LFE}$ and $\boldsymbol{\mathcal{M}}^{\dagger}_{GSCE}$ with two different deformable attention strategies. The white stars denote the query features, and the blue masks denote the regions to which the queries attend.}}
	\label{att_ablation_fig}
\end{figure}

\subsubsection{\textcolor{black}{More analysis}}
\label{more_analysis}

We replace the first conventional layer of $\boldsymbol{\mathcal{M}}_{LFE}$ with a deformable convolutional layer (DCL) \cite{dai2017deformable} and substitute the PPM of $\boldsymbol{\mathcal{M}}_{GSCE}$ with a deformable attention module (DAM) \cite{xia2022vision}. As shown in Fig.~\ref{att_ablation_fig}, in contrast to our proposed (a) $\boldsymbol{\mathcal{M}}_{LFE}$ for local feature extraction and (b) $\boldsymbol{\mathcal{M}}_{GSCE}$ for global self-correlation extraction, the deformed operators (c) and (d) tend to focus on object-relevant regions in a flexible manner, thereby extracting more target-related features. Table \ref{more_analysis_tab} also infers that our proposed SC$^{3}$EF can more effectively estimate dense optical flow maps across visible and thermal modalities by equally extracting local features and indiscriminately establishing global self-correlations, rather than paying more attention to the object-relevant regions.

\section{Conclusion}
\label{conclusion}

In this paper, we propose a novel end-to-end self-correlation and cross-correspondence estimation framework for visible and thermal image registration. The proposed convolution-transformer-based self-correlation extraction and cross-correspondence estimation pipeline is able to utilize intra-modality self-correlation to facilitate inter-modality cross-correspondence estimation. Considering that human observers make use of both local features and global contextual cues to establish correspondences, we design a convolution-based LFE module and a transformer-based GSCE module to extract local representations and global self-correlation within visible and thermal images. Then, the extracted local features and the encoded self-modality correlations are deployed to estimate the cross-correspondences of inter-modality features by our proposed LCCE and GCCE modules in parallel. Finally, the estimated optical flow maps based on local and global cues are merged and hierarchically refined to the full size of the input image. Extensive experimental results demonstrate that our approach outperforms SOTA image registration approaches by large margins, especially for image regions without distinctive features and multispectral images with large displacements. To our best knowledge, this is the first attempt to exploit both local representative features and global contextual cues for RGB-T dense matching by utilizing a convolution-transformer-based architecture.

\begin{table}[tb]
	\tiny
	\renewcommand{\arraystretch}{1.25}
	\centering
	\caption{\textcolor{black}{Quantitative comparison of AEPE and PCK metrics for our proposed SC$^{3}$EF before and after applying $\boldsymbol{\mathcal{M}}^{\dagger}_{LFE}$ and $\boldsymbol{\mathcal{M}}^{\dagger}_{GSCE}$ on the KAIST \cite{hwang2015multispectral} dataset.}}
	\resizebox{0.4\textwidth}{!}{
		\begin{tabular}{ccccc}
			\hline
			\multirow{2}{*}{Metrics} & 
			\multirow{2}{*}{AEPE $\mathbf{\downarrow}$} &
			\multicolumn{3}{c}{PCK ($\%$) $\mathbf{\uparrow}$} \\
			\cline{3-5} & ~  & 1\textit{px} & 3\textit{px} & 5\textit{px} \\
			\hline\hline
			\multicolumn{5}{l}{KAIST \cite{hwang2015multispectral}} \\
			\hline
			SC$^{3}$EF & \textbf{4.65} & \textbf{23.16} & \textbf{62.11} & \textbf{83.61} \\
			SC$^{3}$EF$^{\dagger}$ & \underline{5.02} & \underline{22.09} & \underline{58.51} & \underline{80.48} \\
			\hline\hline
			\multicolumn{5}{l}{RoadScene \cite{xu2020u2fusion}} \\
			\hline
			SC$^{3}$EF & \textbf{9.29} & \textbf{13.90} & \textbf{38.74} & \textbf{57.88} \\
			SC$^{3}$EF$^{\dagger}$ & \underline{10.34} & \underline{11.76} & \underline{33.63} & \underline{51.85} \\
			\hline
	\end{tabular}}
	\label{more_analysis_tab}
\end{table}

\textcolor{black}{Given the significance of degraded RGB-T image registration under challenging lighting and weather conditions, in future work, we plan to investigate the mutual enhancement strategy between cross-modal feature enhancement and cross-correspondence estimation \cite{cho2017visibility} , aiming to boost degraded features by leveraging the complementary characteristics of unaligned RGB-T images, thereby facilitating comprehensive feature encoding for dense correspondence estimation. Besides, based on the practical real-time requirements of autonomous driving and unmanned aerial vehicle (UAV) applications, we will focus on advancing lightweight techniques such as knowledge distillation, model pruning, and neural architecture search to further alleviate the computational burden.}

\bibliographystyle{IEEEtran}
\bibliography{Reference}

\end{document}